\documentclass{article}
\usepackage{caption}
\usepackage{spconf,amsmath,graphicx}
\usepackage{amsthm}
\usepackage{amsfonts}
\usepackage{float}
\usepackage{subfig}
\theoremstyle{plain}
\usepackage{color}

\newtheorem{Definition}{Definition}
\newtheorem{Theorem}{Theorem}
\newtheorem{Lemma}{Lemma}

\title{Learned ISTA with Error-based Thresholding for Adaptive Sparse Coding}
\name{Ziang Li$^{1}$ \qquad Kailun Wu$^{2}$ \qquad Yiwen Guo $^{2\ast}$ \qquad Changshui Zhang$^{1\ast}$\thanks{$\ast$ Corresponding author. This work is funded by the Natural Science Fundation of China (NSFC. No. 62176132).}}
\address{$^{1}$ Institute for Artificial Intelligence, Tsinghua University (THUAI), \\Beijing National Research Center for Information Science and Technology (BNRist), \\Department of Automation, Tsinghua University, Beijing, P.R.China\\ $^{2}$ Independent Researcher}

\begin{document}
%
\maketitle
\begin{abstract}
  Drawing on theoretical insights, we advocate an error-based thresholding (EBT) mechanism for learned ISTA (LISTA), which utilizes a function of the layer-wise reconstruction error to suggest a specific threshold for each observation in the shrinkage function of each layer. We show that the proposed EBT mechanism well disentangles the learnable parameters in the shrinkage functions from the reconstruction errors, endowing the obtained models with improved adaptivity to possible data variations. With rigorous analyses, we further show that the proposed EBT also leads to a faster convergence on the basis of LISTA or its variants, in addition to its higher adaptivity. Extensive experimental results confirm our theoretical analyses and verify the effectiveness of our methods.
\end{abstract}
\section{Introduction}
\label{intr}
The core problem of sparse coding is to deduce the high-dimensional sparse code from the low-dimensional observation. The basic assumption can be formulated as $y=Ax^\star+\varepsilon$,
where $y\in \mathbb{R}^{m}$ is the observation corrupted by the inevitable noise $\varepsilon \in \mathbb{R}^{m}$, $x^\star\in \mathbb{R}^{n}$ is the sparse code to be estimated, and $A\in \mathbb{R}^{m\times n}$ is an over-complete dictionary matrix. The main challenge to estimate $x$ is its ill-posed nature because of over-complete modeling, i.e., $m<n$. A possible solution is to solve the LASSO problem formulated as $\min_{x}\|y-Ax\|_2+\lambda\|x\|_1$
using the iterative shrinking thresholding algorithm (ISTA)~\cite{daubechies2004iterative}. To achieve faster convergence, Learned ISTA (LISTA) \cite{gregor2010learning} was then proposed in the deep learning era, in which the architecture of the deep neural network (DNN) followed the iterative process of ISTA. The thresholding mechanism was then modified to the learned thresholds in shrinkage functions of DNNs.

Yet, LISTA and many other deep networks based on LISTA~\cite{chen2018theoretical,liu2019alista,ablin2019learning,Wu2020gated,aberdam2021ada,zhou2018sc2net} suffer the issue that the thresholds were shared among all training samples, which means they lacked adaptability to the variety of training samples and robustness to outliers. Also, it leads to poor generalization to test data with a different distribution (or sparsity) from the training data. To address the above issues, we propose an error-based thresholding (EBT) mechanism of LISTA-based models to improve their adaptivity. EBT introduces a function of the evolving estimation error to provide each threshold in the shrinkage functions in the model. It has no extra learnable parameter compared with original LISTA-based models, yet shows significantly better performance. 

\section{Preliminary Knowledge}
\label{bg}
The update rule of ISTA is
\begin{equation}
\label{eq:ISTA}
x^{(t+1)}=\mathrm{sh}_{\lambda/\gamma}((I-A^TA/\gamma)x^{(t)}+A^Ty/\gamma), \quad  \forall t\geq 0,
\end{equation}
where $\mathrm{sh}_b(x) = \mathrm{sign}(x)(|x|-b)_+$ is a shrinkage function with a threshold $b\geq 0$ and $(\cdot)_+ = \max\{0,\cdot\}$,  $\gamma$ is the maximal eigenvalue of the symmetric matrix $A^TA$. LISTA kept the update rule of ISTA but learned parameters via end-to-end training. Its inference process can be formulated as 
\begin{equation}
\label{LISTA_origin}
x^{(t+1)}=\mathrm{sh}_{b^{(t)}}(W^{(t)}x^{(t)}+U^{(t)}y),
\end{equation}
where $\Theta=\{W^{(t)},U^{(t)},b^{(t)}\}_{t=0,\dots,d}$ is a set of learnable parameters. Specifically, $b^{(t)}$ is the layer-wise learnable threshold shared among all samples, which is of our particular interest in this paper. LISTA has been proved to converge linearly with partial weight coupling~\cite{chen2018theoretical}, i.e., $W^{(t)} = I - U^{(t)}A$, thus Eq.~(\ref{LISTA_origin}) can be written as
\begin{equation} 
\label{LISTA}
x^{(t+1)}=\mathrm{sh}_{b^{(t)}}((I-U^{(t)}A)x^{(t)}+U^{(t)}y).
\end{equation}
Support selection was further introduced to modify LISTA~\cite{chen2018theoretical}. We select entries with largest $p^{(t)}$ (in percentage) magnitudes in vector $x$ as support set $S_{p^{(t)}}$. The entries in support set do not pass the thresholding and keep them values unchanged. Formally, we define $\mathrm{shp}_{(b^{(t)},p^{(t)})}$ as the thresholding operator with support selection, the update rule is then formulated as
\begin{equation}
\label{listacpss}
x^{(t+1)}=\mathrm{shp}_{(b^{(t)},p^{(t)})}((I-U^{(t)}A)x^{(t)}+U^{(t)}y),
\end{equation}
where $p^{(t)}$ is a hyper-parameter and it increases from lower layers to higher layers. $p^{(t)}$ can be set as $p^{(t)}=\min(p\cdot t, p_{max})$, where $p$ is the positive scalar and $p_{max}$ is the upper bound of the percentage of non-zero elements in $x$. LISTA with support selection is proved to achieve faster convergence in comparison with LISTA~\cite{chen2018theoretical}. 

According to the convergence analyses of LISTA and its variants~\cite{chen2018theoretical,liu2019alista,ablin2019learning,Wu2020gated}, the following equality should hold for the threshold at each layer to ensure linear convergence:
\begin{equation}
\label{thresh}
b^{(t)} = \mu(A)\sup_{x^\star\in \mathcal{S}}\|x^{(t)}-x^\star\|_\phi
\end{equation}
where $\mathcal{S}$ is the training set, $\mu(A)$ is the generalized mutual coherence coefficient of the dictionary matrix $A\in \mathbb{R}^{m\times n}$, and $\phi$ represents the type of the norm which is commonly set as $1$ or $2$. Note that $\mu(A)$ is a crucial term in this paper, here we formally give its definition together with the definition of $\mathcal{W}(A)$ as follows:

\begin{Definition}
	\label{def1}
	\rm{\cite{liu2019alista}}
    For a matrix $A$, we let $A_{i,:}$ represents the $i$-th row of $A$ and $A_{:,j}$ represents the $j$-th column of $A$. The generalized mutual coherence coefficient of $A$ is $\mu(A)=\inf_{W\in \mathbb{R}^{n\times m},W_{i,:}A_{:,i}=1}\max_{i\neq j}{(W_{i,:}A_{:,j})}$. In addition, we let $\mathcal{W}(A)$ denotes the set of the matrices which can achieve $\mu(A)$, which means $\mathcal{W}(A) = \{W\in \mathbb{R}^{n\times m}|\max_{i\neq j}{(W_{i,:}A_{:,j})}=\mu(A),W_{i,:}A_{:,i}=1,\forall i\}$.
\end{Definition}


\section{Methods}
\label{methods}
In LISTA and many of its variants, the threshold $b^{(t)}$ is commonly treated as a learnable parameter. As demonstrated in Eq.~(\ref{thresh}), $b^{(t)}$ should be proportional to the upper bound of the estimation error of the $t$-th layer in the noiseless case to ensure fast convergence~\cite{chen2018theoretical,liu2019alista,ablin2019learning,Wu2020gated}. Thus, some outliers or extreme training samples greatly influence the value of $b^{(t)}$, making the obtained threshold not fit the majority of the data.

In order to solve this problem, we propose to disentangle the reconstruction error term from the learnable part of the threshold and introduce adaptive thresholds for LISTA and related networks. We attempt to rewrite the threshold at the $t$-th layer as something like $b^{(t)} = \rho^{(t)} \|x^{(t)}-x^\star\|_\phi,$ where $\rho^{(t)}$ is a layer-specific learnable parameter. However, the ground-truth $x^\star$ is actually unknown for the inference process. Therefore, we need to find an alternative formulation.
Notice that in the noiseless case, it holds that $Ax^{(t)}-y=A(x^{(t)}-x^\star)$. Also, we know $U^{(t)}\in \mathcal{W}(A)$ is desirable according to prior works~\cite{chen2018theoretical,liu2019alista,ablin2019learning,Wu2020gated}, which means $U^{(t)}A$ approximates the identity matrix. Thus, we propose our EBT-LISTA, which is formulated as  
\begin{equation}
\label{thresh_2}
\begin{aligned}
x^{(t+1)}&=\mathrm{sh}_{b^{(t)}}((I-U^{(t)}A)x^{(t)}+U^{(t)}y),\\
b^{(t)}&=\rho^{(t)}\|U^{(t)}(Ax^{(t)}-y)\|_\phi.
\end{aligned}
\end{equation}
Note that only $\rho^{(t)}$ and $U^{(t)}$ are learnable parameters in the above formulation, thus our EBT-LISTA actually introduces no extra parameters compared with the original LISTA.

We can also apply our EBT mechanism to LISTA with support selection~\cite{chen2018theoretical} by keeping support selection operation and replacing the fixed threshold, which is formulated as 
\begin{equation}
\label{eq2}
\begin{aligned}
x^{(t+1)}&=\mathrm{shp}_{(b^{(t)},p^{(t)})}((I-U^{(t)}A)x^{(t)}+U^{(t)}y),\\
b^{(t)}&=\rho^{(t)}\|U^{(t)}(Ax^{(t)}-y)\|_\phi.
\end{aligned}
\end{equation}
\section{Theoretical Analysis}
\label{TA}
In this section, we provide convergence analyses for LISTA and LISTA with support selection. Proof of all our theoretical results can be found in the appendices\footnote{Appendix can be seen on https://arxiv.org/abs/2112.10985.}. We focus on the noiseless case and the main results are obtained under some assumptions of the ground-truth sparse code. 
Specifically, we here assume that the ground-truth sparse vector $x^\star$ is sampled from the distribution $\gamma(B,s)$, i.e., the number of its nonzero elements follow a uniform distribution $U(0,s)$ and the magnitude of its nonzero elements follow an arbitrary distribution in $[-B,B]$. We also assume that $s$ is sufficiently small in our theoretical analysis for error-based thresholding, which means $\mu(A)(2s-1) < 1$ to be exact. Note that similar assumptions can also be found in relative works~\cite{chen2018theoretical,liu2019alista,Wu2020gated,aberdam2021ada}. 

\subsection{EBT mechanism on LISTA}
Let us first recall the convergence of LISTA and discuss how our EBT improves LISTA in accelerating convergence. 
\begin{Lemma} 
	\label{lemma1}
	\rm{\cite{chen2018theoretical}}
	For LISTA formulated in Eq.~(\ref{LISTA}), $x^\star$ is sampled from $\gamma(B,s)$. If $s$ is small such that $\mu(A)(2s-1)<1$, $U^{(t)} \in \mathcal{W}(A)$, and $b^{(t)} = \mu(A)\sup_{x^\star}\|x^{(t)}-x^\star\|_1$,
	the estimation $x^{(t)}$ at the $t$-th layer of LISTA satisfies $$\|x^{(t)}-x^\star\|_2\leq sB\exp{(c_0t)},$$ where $c_0 = \log((2s-1)\mu(A)) < 0$.
\end{Lemma}
\begin{Theorem}
	\label{thm1}
	\rm{\textbf{(Convergence of EBT-LISTA)}}
	For EBT-LISTA formulated in Eq.~(\ref{thresh_2}) where $\phi=1$, $x^\star$ is sampled from $\gamma(B,s)$. If $s$ is small such that $\mu(A)(2s-1) < 1$, $U^{(t)} \in \mathcal{W}(A)$ and $\rho^{(t)} = \frac{\mu(A)}{1-\mu(A)s}$,
	the estimation $x^{(t)}$ at the $t$-th layer satisfies $$\|x^{(t)}-x^\star\|_2\leq q_0\exp{(c_1t)},$$ where $q_0<sB$ and $c_1<c_0$ hold with the probability of $1 - \mu(A)s$.
\end{Theorem}
Compared Theorem \ref{thm1} with Lemma~\ref{lemma1}, we know that EBT-LISTA converges similarly as the original LISTA. The convergence rate is probably faster and the reconstruction error is probably lower, with $c_1 < c_0$ and $q_0<sB$. Since $s$ is assumed to be small such that $\mu(A)(2s-1) < 1$, which means $\mu(A)s<0.5$, indicating our EBT achieves superiority in a higher probability. Moreover, in an extremely sparse scenario with $s$ being sufficiently small such that $\mu(A)s \ll 1$, the probability of achieving the superiority is relatively high in theory. Under this circumstance, the desired threshold in EBT-LISTA should be $\mu(A)$ and it is disentangled with the reconstruction error, unlike the desired threshold in original LISTA, i.e., $\mu(A)\sup_{x^\star}\|x^{(t)}-x^\star\|_1$.

\subsection{EBT mechanism on LISTA with support selection}
There exist many variants of LISTA, and in this subsection, we choose LISTA with support set selection as an example to show how our EBT improves it in theory. 
\begin{Lemma}
	\label{lemma2}
	\rm{\textbf{(Convergence of LISTA with support selection)}}
	For LISTA with support selection formulated in Eq.~(\ref{listacpss}), $x^\star$ is sampled from $\gamma(B,s)$. If $s$ is small such that $\mu(A)(2s-1)<1$, $U^{(t)} \in \mathcal{W}(A)$ and $b^{(t)} = \mu(A)\sup_{x^\star}\|x^{(t)}-x^\star\|_1$,
	there actually exist two convergence phases. In the first phase, i.e., $t\leq t_0$, the $t$-th layer estimation $x^{(t)}$ satisfies $$\|x^{(t)}-x^\star\|_2\leq sB\exp{(c_2t)},$$ where $c_2 \leq \log((2s-1)\mu(A))$.
	In the second phase, i.e., $t> t_0$, the estimation $x^{(t)}$ satisfies $$\|x^{(t)}-x^\star\|_2\leq C \|x^{(t-1)}-x^\star\|_2,$$ where $C \leq s\mu(A)$. 
\end{Lemma}
\begin{Theorem}
	\label{thm2}
	\rm{\textbf{(Convergence of EBT-LISTA with support selection)}}
	For EBT-LISTA with support selection and $\phi=1$, $x^\star$ is sampled from $\gamma(B,s)$. If $s$ is small such that $\mu(A)(2s-1) < 1$, $U^{(t)} \in \mathcal{W}(A)$, and $\rho^{(t)} = \frac{\mu(A)}{1-\mu(A)s}$,
	there exist two convergence phases. In the first phase, i.e., $t\leq t_1$, the $t$-th layer estimation $x^{(t)}$ satisfies $$\|x^{(t)}-x^\star\|_2\leq q_1\exp{(c_3t)},$$ where $c_3<c_2$, $q_1 < sB$ and $t_1<t_0$ hold with a probability of $1 - \mu(A)s$.
	In the second phase, i.e., $t> t_1$, the estimation $x^{(t)}$ satisfies $$\|x^{(t)}-x^\star\|_2\leq C \|x^{(t-1)}-x^\star\|_2,$$ where $C \leq s\mu(A)$. 
\end{Theorem}
Lemma~\ref{lemma2} and Theorem~\ref{thm2} show that when powered with support selection, LISTA shows two different convergence phases. The earlier phase is generally slower and the later phase is faster. After incorporating EBT, the model processes the same rate of convergence in the second phase. While in the first phase, our EBT leads to faster convergence, and it thus gets in the second phase faster, showing the effectiveness of our EBT in LISTA with support selection in theory.

\section{Experiments}

\label{ep}
We conduct experiments on both synthetic data and real data to validate our theorem and test our methods. $p_b$ is set as the sparsity (i.e., the probability of any of its entries be zero) of $x^\star$. Other experimental settings can be found in appendices.


\begin{figure*}[t]
	\begin{minipage}{1.0\linewidth}
		\centering
		\vskip 0.05in
		\subfloat[$p_b$ changes from 0.8 to 0.99]{\label{fig:20to1}
			\includegraphics[width=0.29\linewidth]{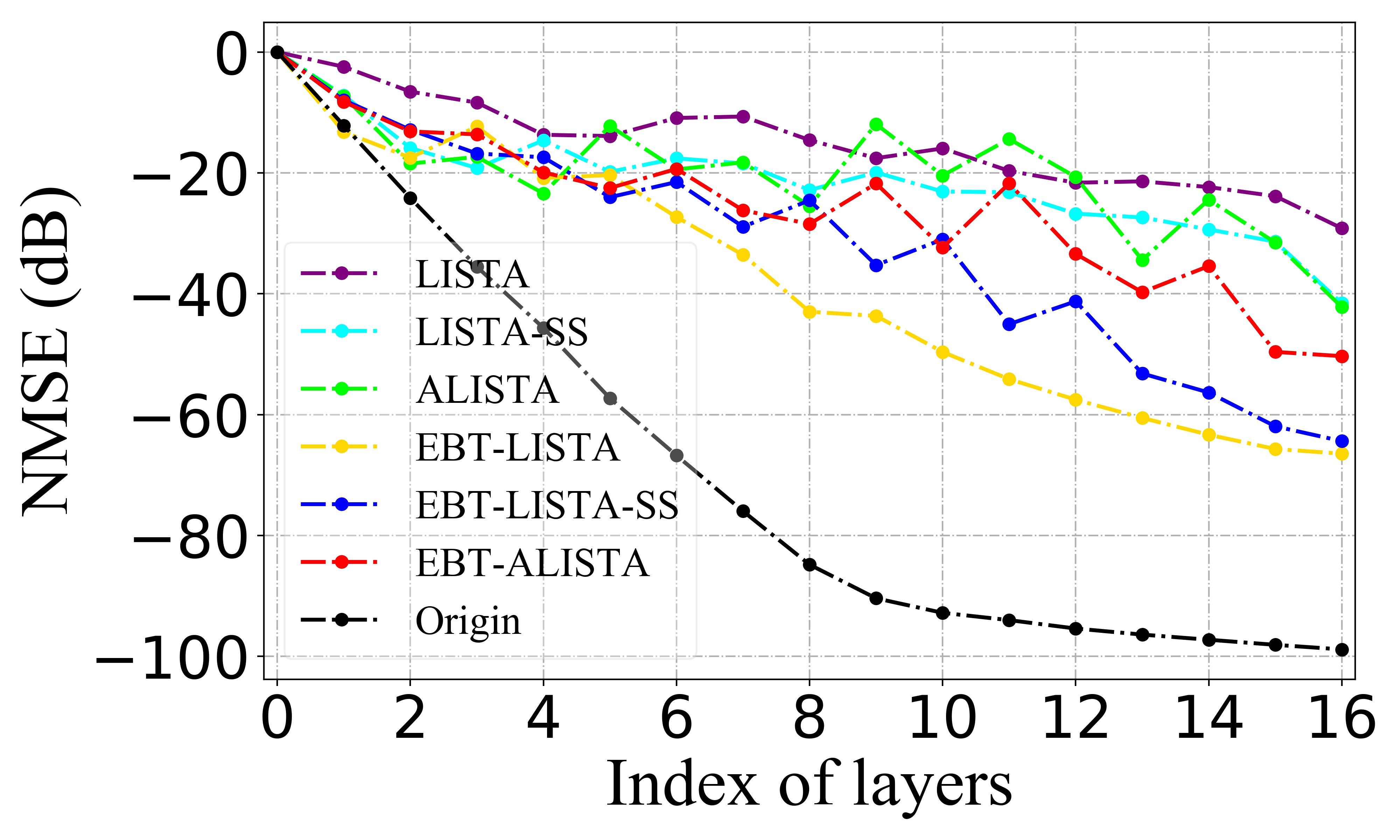}}	
		\hskip 0.1in
		\subfloat[$p_b$ changes from 0.8 to 0.9]{\label{fig:20to10}
			\includegraphics[width=0.29\linewidth]{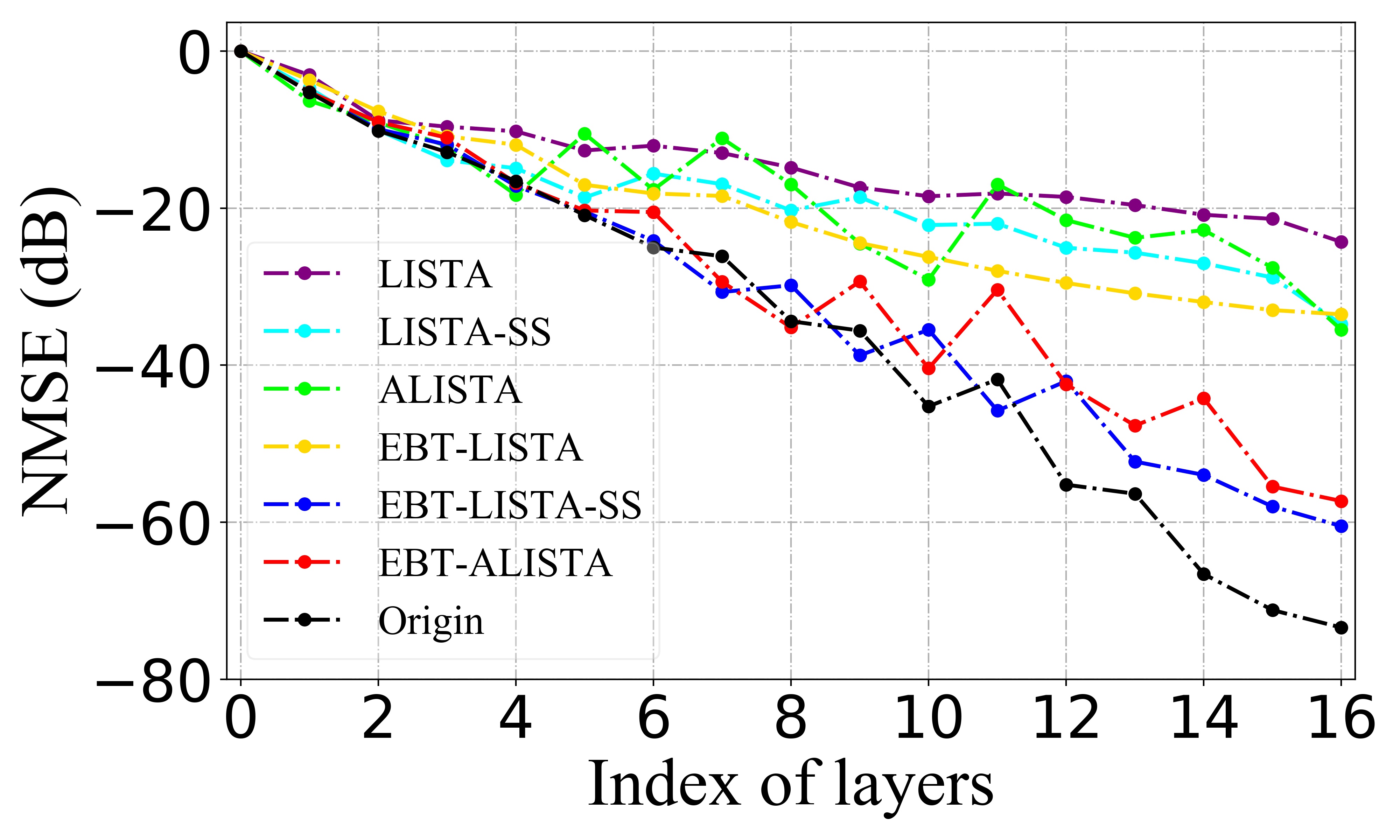}}
		\hskip 0.1in
		\subfloat[$p_b$ changes from 0.9 to 0.8]{\label{fig:10to20}	
			\includegraphics[width=0.29\linewidth]{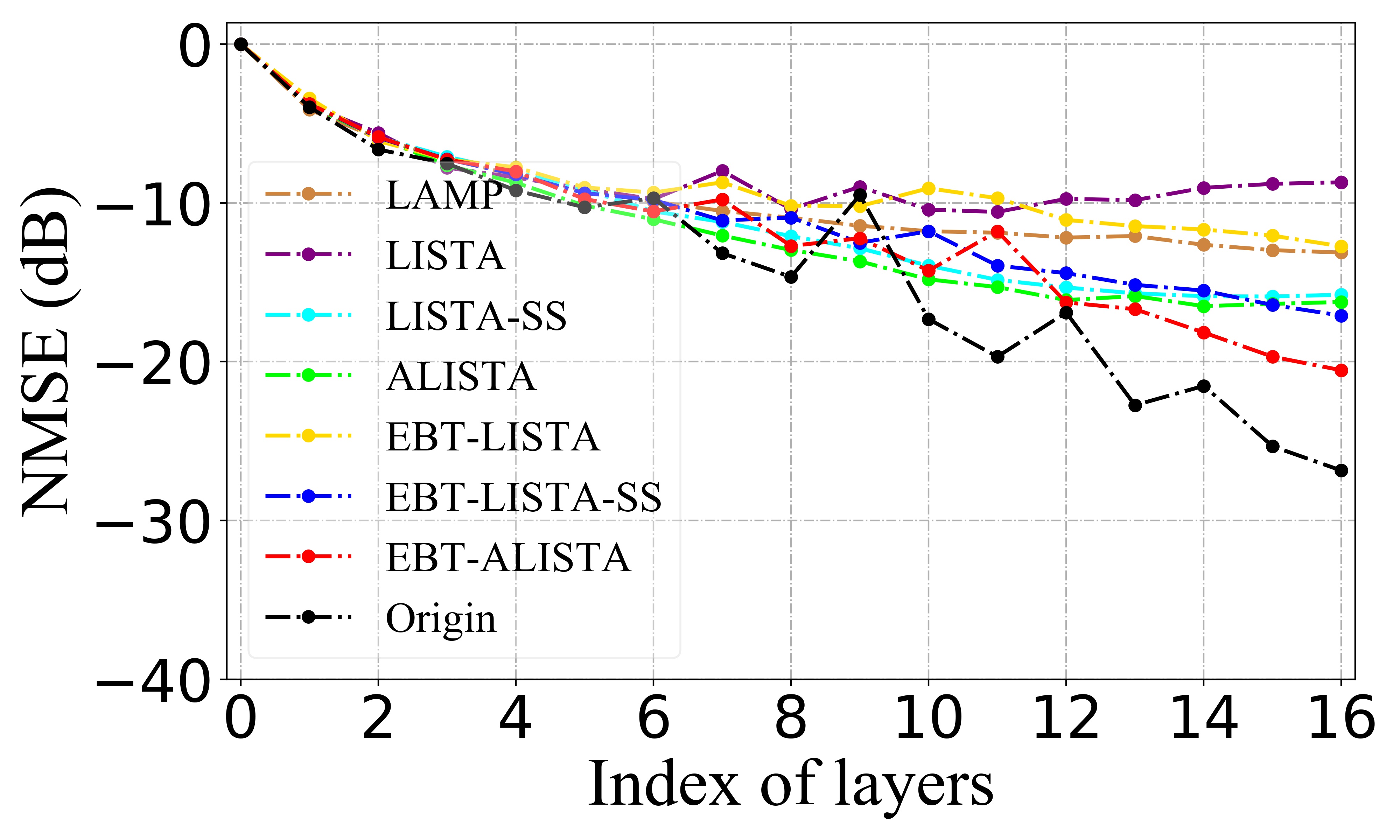}}
		\vskip -0.1in
		\caption{NMSE of different models when the test sparsity is different from the training sparsity.  We use ``Origin'' to indicate the optimal scenario where the LISTA models are trained on exactly the same sparsity as that of the test data.}\vskip 0.05in
		\label{fig:pb}
	\end{minipage}
	\begin{minipage}{1.0\linewidth}
		\centering
		\subfloat[$p_b$=0.95]{\label{fig:P05}
			\includegraphics[width=0.29\linewidth]{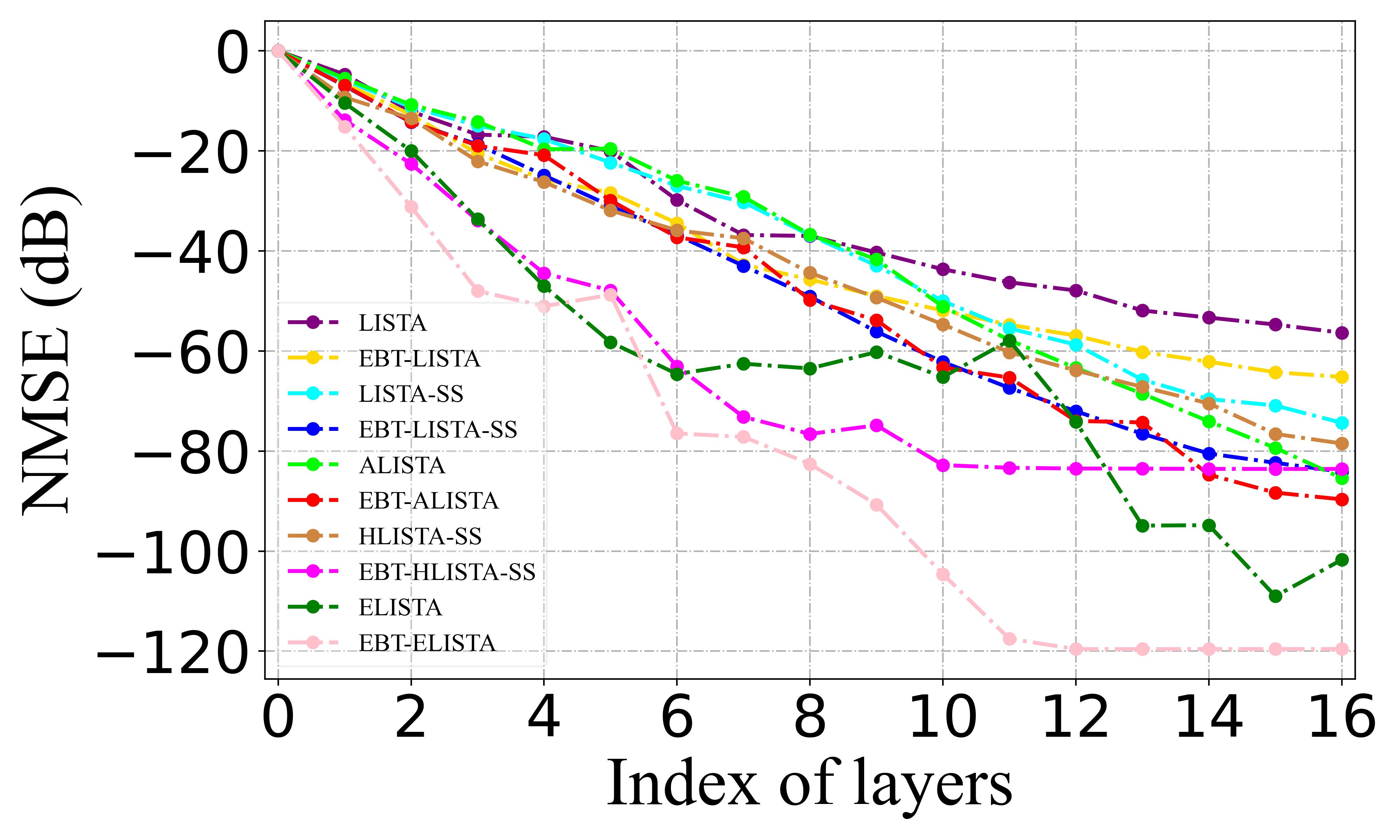}}
		\hskip 0.1in
		\subfloat[$p_b$=0.9]{\label{fig:inf}
			\includegraphics[width=0.29\linewidth]{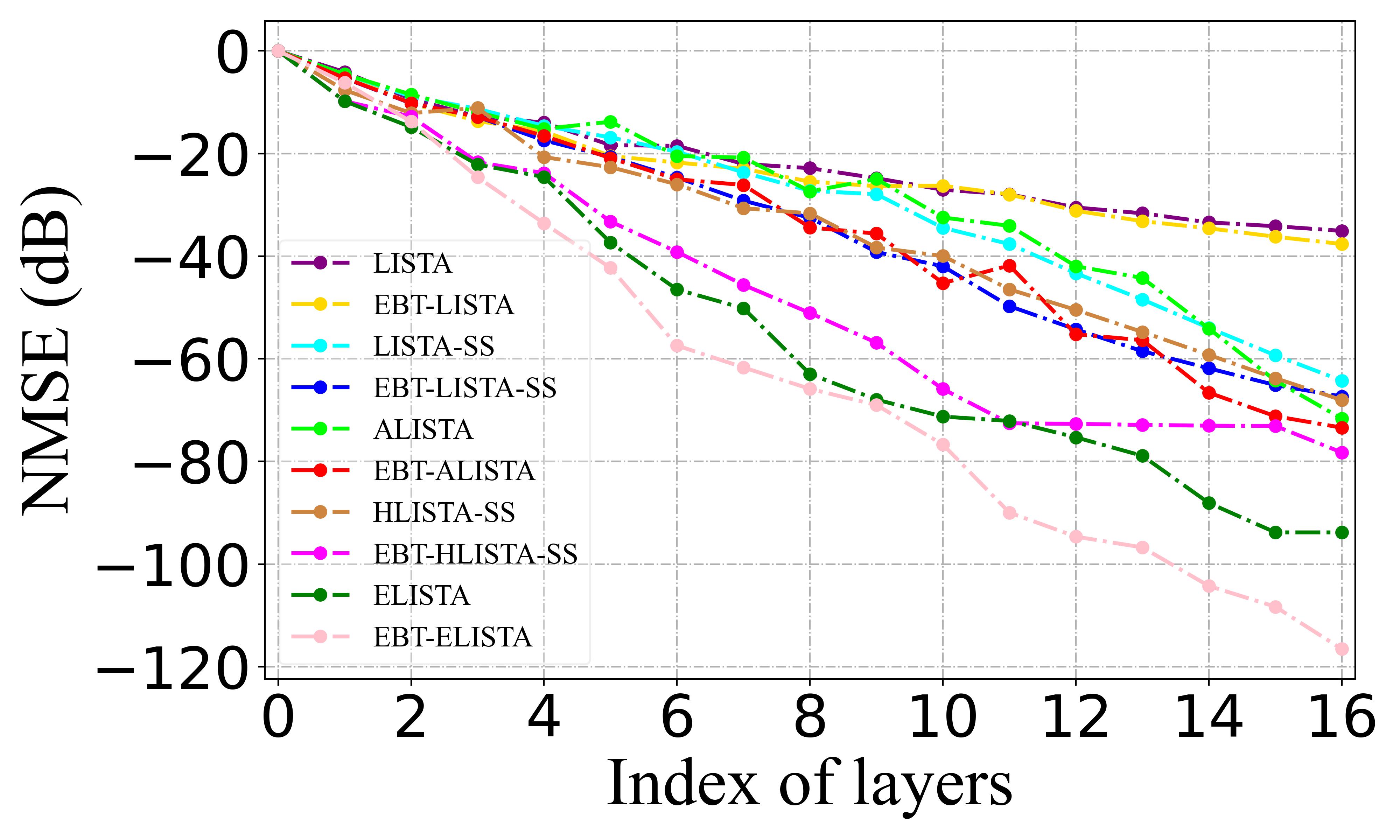}}
		\hskip 0.1in
		\subfloat[$p_b$=0.8]{\label{fig:P20}
			\includegraphics[width=0.29\linewidth]{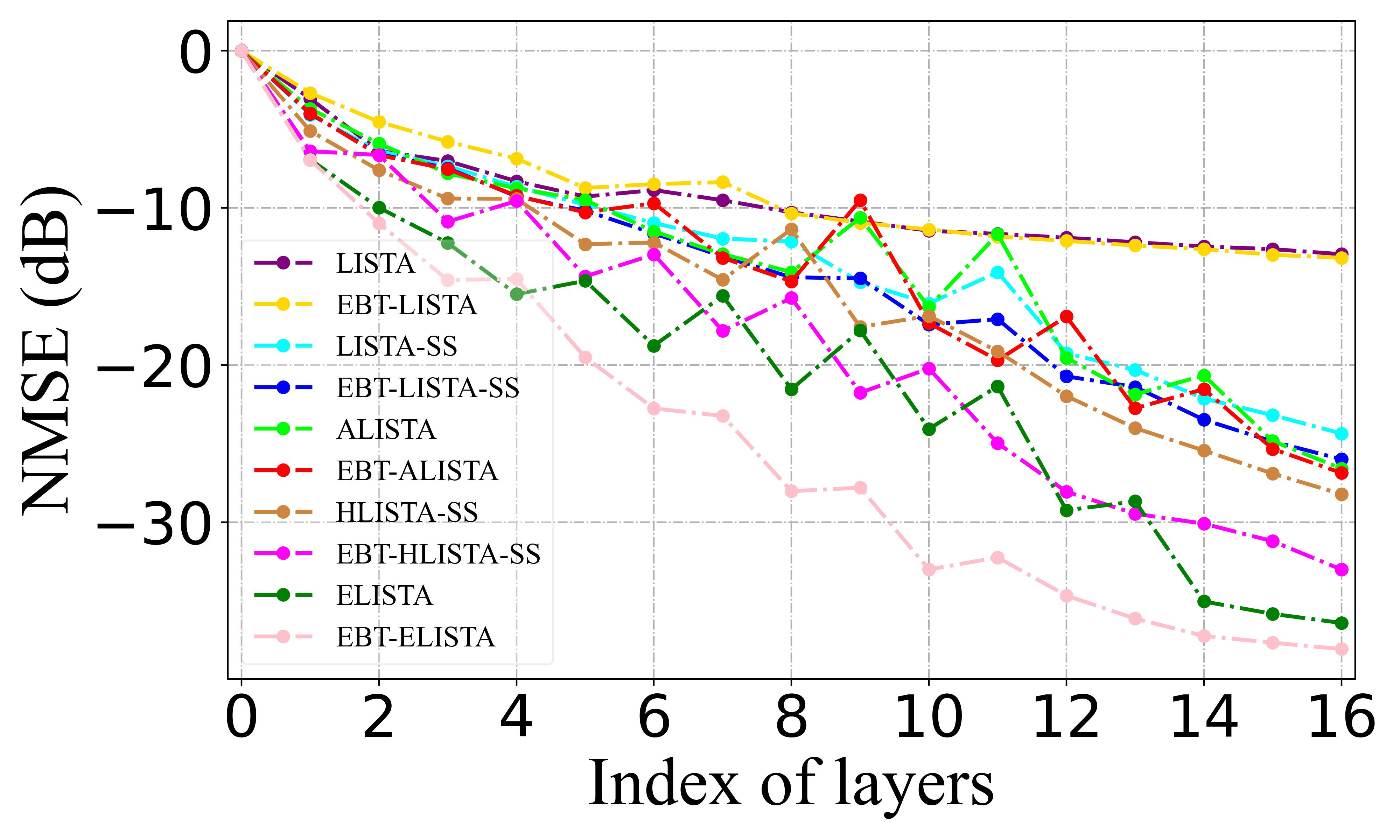}}
            \vskip -0.1in
		\caption{NMSE of different models under different sparsity. When combined with our EBT, faster convergence is obtained. } 
		\label{fig:settings}
	\end{minipage} \vskip -0.1in
\end{figure*}

\subsection{Simulation Experiments} 

\begin{figure}[ht]
	\label{fig:function}
	\centering
	\subfloat[LISTA]{\label{fig:valid1}	
		\includegraphics[width=0.45\linewidth]{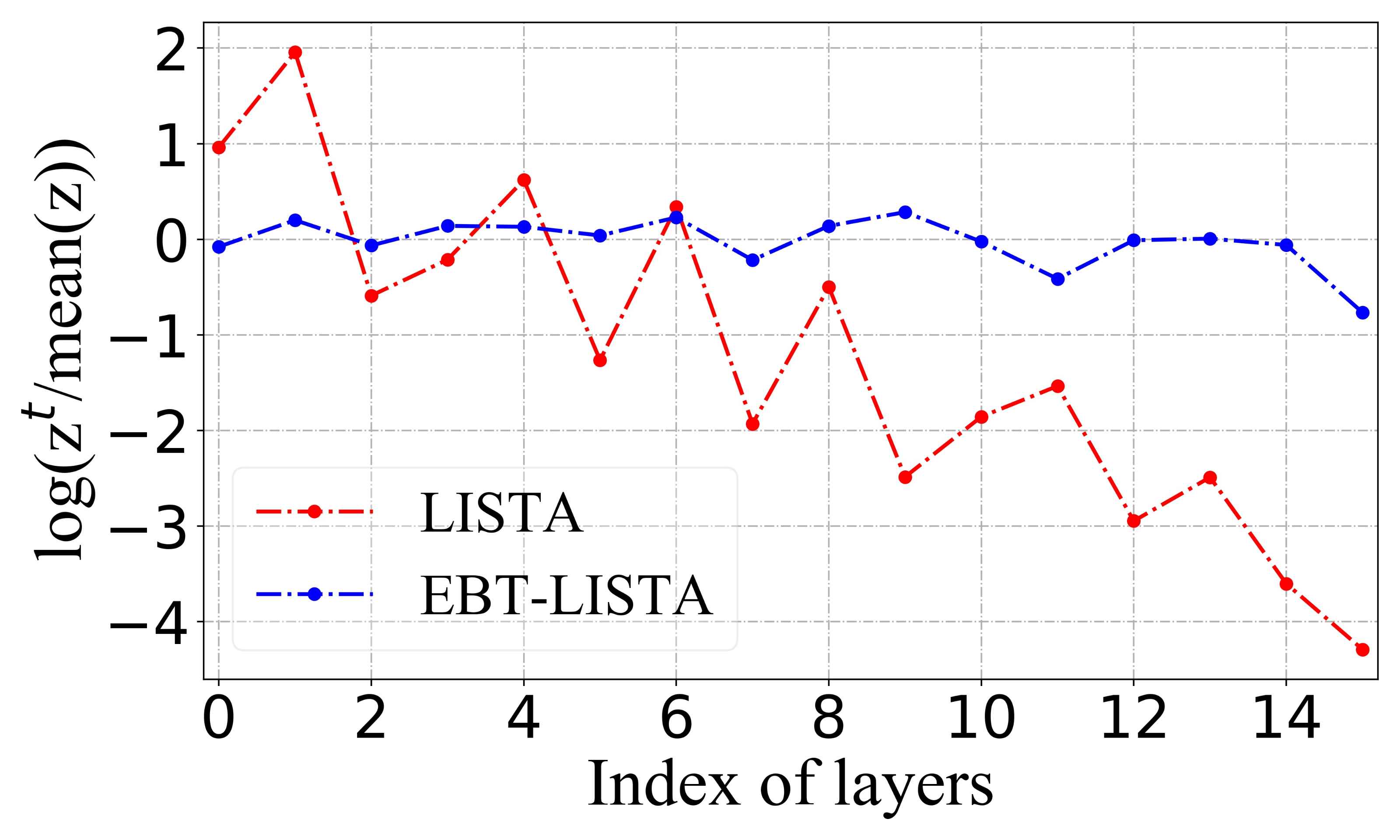}}
	\hskip 0.1in
	\subfloat[LISTA-SS]{\label{fig:valid2}	
		\includegraphics[width=0.45\linewidth]{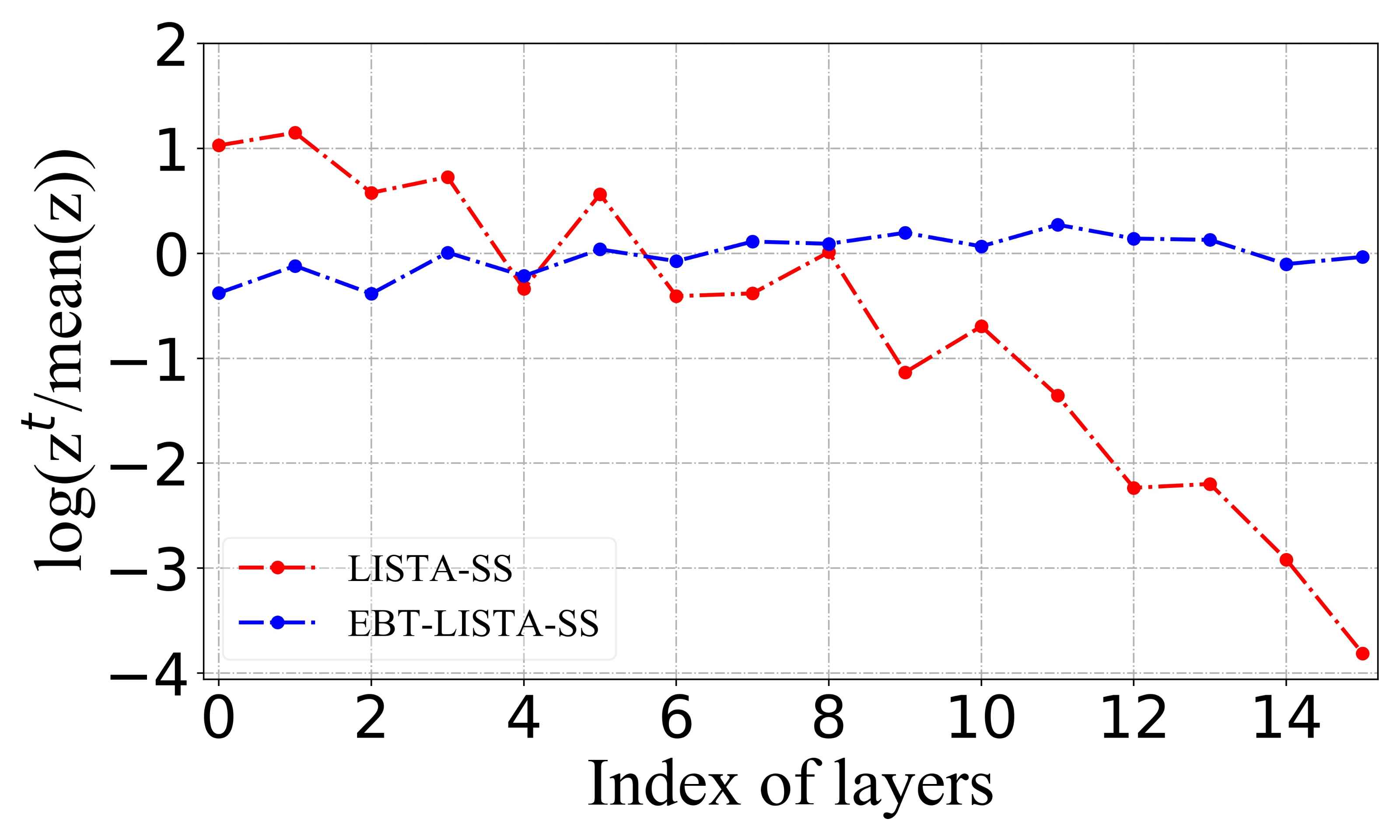}}
        \vskip -0.1in
	\caption{Disentanglement of the reconstruction error and learnable parameters in our EBT. $z^{(t)}$ here indicates $\rho^{(t)}$ and $b^{(t)}$ for networks with or without EBT, respectively.} 
	\vskip -0.2in
\end{figure}

\hspace{1em} \textbf{Disentanglement.} First, we would like to compare the learned parameters (i.e., $b^{(t)}$ and $\rho^{(t)}$) for the thresholds in both LISTA and our EBT-LISTA. Figure \ref{fig:valid1} shows how the learned parameters (in a logarithmic coordinate) vary across layers in LISTA and EBT-LISTA. Note that the mean values are removed to align the range of the parameters of different models on the same y-axis. It can be seen that the obtained values for the parameter in EBT-LISTA do not change much from lower layers to higher layers, while the reconstruction errors in fact decrease. By contrast, the obtained threshold values in LISTA vary a lot across layers. Such results imply that the optimal thresholds in EBT-LISTA are indeed independent to (or say disentangled from) the reconstruction error, which well confirms our theoretical result in Theorem \ref{thm1}. Similar observations can also be made on LISTA-SS (i.e., LISTA with support selection) and our EBT-LISTA-SS. 

\textbf{Adaptivity to unknown sparsity.}
As have been mentioned, in some practical scenarios, there may exist an obvious gap between the training and test data distribution, or we may not know the distribution of real test data and have to train on synthesized data based on the guess of the test distribution. Under such circumstances, it is of importance to consider the adaptivity/generalization of the sparse coding model (trained on a specific data distribution or with a specific sparsity and the test data sampled from different distributions with different sparsity).
To evaluate in such scenario, we let the test sparsity be different from the training sparsity. Figure \ref{fig:pb} shows the results in three different test settings (the performance is evaluated by the normalized mean squared error (NMSE) in decibels (dB)).
The black curves represent the optimal model when LISTA is trained on exactly the same sparsity as that of the test data. It can be seen that our EBT has huge advantages in such a practical scenario where the adaptivity to un-trained sparsity is required, and the performance gain is larger when the distribution shift between training and test is larger (cf. purple line and yellow line in Figures~\ref{fig:20to1} and~\ref{fig:20to10}).

\textbf{Combination with other methods.}
As previously mentioned, our EBT can be combined with many prior efforts (in addition to LISTA, LISTA-SS, and ALISTA). Here we will show its effectiveness with ELISTA~\cite{li2021learned} and HLISTA with support selection (HLISTA-SS)~\cite{zheng2022hybrid}. 
For the methods with support selection (i.e., LISTA-SS, EBT-LISTA-SS, ALISTA, EBT-ALISTA, HLISTA-SS,and EBT-HLISTA-SS), we adopt $p=0.6$ and $p_{max}=6.5$ for $p_b=0.95 $. We let $p$ and $p_{max}$ be 1.2 and 13.0 for $p_b=0.9$, and let them be 1.5 and 16.25 for $p_b=0.8$. Figure \ref{fig:settings} demonstrates some of the results in different sparsity settings, while more results including different noise levels and different condition numbers can be found in the appendices. In all settings, we can see that our EBT leads to significantly faster convergence. In addition, the superiority of our EBT-based models is more significant with a larger $p_b$ for which the assumption of $s$ is more likely to hold. 

\subsection{Photometric Stereo Analysis}
We also consider a practical sparse coding task: photometric stereo analysis~\cite{ikehata2012robust}. The task solves the problem of estimating the normal direction of a Lambertian surface, given $q$ observations under different light directions $L\in\mathbb R^{q\times 3}$. Note that the noise is sparse, thus we estimate the noise first, and then calculate the desired normal direction.

{\renewcommand\arraystretch{1.2}
	\begin{table}[htb]
		\vskip -0.1in
		\caption{Mean error (in degree) with different number of observations and different test sparsity.} \vskip -0.15in
		\label{table:lb}
		\begin{center}
			\begin{small}
				\setlength{\tabcolsep}{1.5mm}{\begin{tabular}{ccc|cc}
						\hline
						$p_e$ & $q$  & $l_1$ & LISTA-SS & EBT-LISTA-SS\\
						\hline
						&15&0.678&$5.50\times10^{-2}$&$4.09\times10^{-2}$\\
						0.8&25&0.408&$7.48\times10^{-3}$&$3.17\times10^{-3}$\\
						&35&0.336&$1.89\times10^{-3}$&$5.95\times10^{-4}$\\
						\hline
						&15&0.232&$6.67\times10^{-3}$&$2.57\times10^{-3}$\\
						0.9&25&0.145&$1.33\times10^{-3}$&$1.64\times10^{-4}$\\
						&35&0.088&$2.93\times10^{-4}$&$4.91\times10^{-5}$\\
						\hline
				\end{tabular}} 
			\end{small}
		\end{center}
		\vskip -0.27in
\end{table}}

In this experiment, we mainly follow the settings in ~\cite{xin2016maximal} and ~\cite{Wu2020gated}. We use the same bunny picture for evaluation and $L$ is also randomly selected from the hemispherical surface. We set the number of observations $q$ to be $15$, $25$, and $35$, and training sparsity is $p_{t}=0.8$. The final performance is evaluated by calculating the average angle between the estimated normal vector and the ground-truth normal vector (in degree). Since the distribution of the noise is generally unknown in practice, the adaptivity is of greater importance for this task. We use two test settings for evaluating different models, in which the sparsity of the noise in test data (i.e., $p_e$) is set as 0.8 and 0.9, respectively. 
We compare EBT-LISTA-SS, LISTA-SS, and a conventional methods, \emph{i.e.,} least 1-norm ($l_1$). 
Results in Table~\ref{table:lb} 
show that EBT-LISTA-SS outperforms in the concerned settings, which means our EBT-based network can be more effective in this task. Note that the advantage is more remarkable when $p_e = 0.9$, i.e., $p_e\neq p_t$, which means EBT-based network has better adaptivity than the original LISTA-based networks.
\section{Conclusion}
\label{conclusion}
In this paper, we have studied the thresholds in the shrinkage functions of LISTA. We have proposed a novel mechanism called EBT which well disentangles the learnable parameter in the shrinkage function on each layer of LISTA from its layer-wise reconstruction error. We have proved theoretically that, in combination with LISTA or its existing variants, our EBT mechanism leads to faster convergence and achieves superior final sparse coding performance. Also, we have shown that the EBT mechanism endows deep unfolding models with higher adaptivity to different observations with a variety of sparsity. Our experiments on both synthetic data and real data have testified the effectiveness of our EBT, especially when the distributions of the test and training data are different.

\bibliographystyle{IEEEbib}
\bibliography{refs}

\begin{thebibliography}{10}

\bibitem{daubechies2004iterative}
Ingrid Daubechies, Michel Defrise, and Christine De~Mol,
\newblock ``An iterative thresholding algorithm for linear inverse problems
  with a sparsity constraint,''
\newblock {\em Communications on Pure and Applied Mathematics: A Journal Issued
  by the Courant Institute of Mathematical Sciences}, vol. 57, no. 11, pp.
  1413--1457, 2004.

\bibitem{gregor2010learning}
Karol Gregor and Yann LeCun,
\newblock ``Learning fast approximations of sparse coding,''
\newblock in {\em Proceedings of the 27th International Conference on
  International Conference on Machine Learning}. Omnipress, 2010, pp. 399--406.

\bibitem{chen2018theoretical}
Xiaohan Chen, Jialin Liu, Zhangyang Wang, and Wotao Yin,
\newblock ``Theoretical linear convergence of unfolded ista and its practical
  weights and thresholds,''
\newblock in {\em Advances in Neural Information Processing Systems}, 2018, pp.
  9061--9071.

\bibitem{liu2019alista}
Jialin Liu, Xiaohan Chen, Zhangyang Wang, and Wotao Yin,
\newblock ``Alista: Analytic weights are as good as learned weights in lista,''
\newblock in {\em International Conference on Learning Representations (ICLR)},
  2019.

\bibitem{ablin2019learning}
Pierre Ablin, Thomas Moreau, Mathurin Massias, and Alexandre Gramfort,
\newblock ``Learning step sizes for unfolded sparse coding,''
\newblock in {\em Advances in Neural Information Processing Systems}, 2019, pp.
  13100--13110.

\bibitem{Wu2020gated}
Kailun Wu, Yiwen Guo, Ziang Li, and Changshui Zhang,
\newblock ``Sparse coding with gated learned ista,''
\newblock in {\em Proceedings of the International Conference on Learning
  Representations}, 2020.

\bibitem{aberdam2021ada}
Aviad Aberdam, Alona Golts, and Michael Elad,
\newblock ``Ada-lista: Learned solvers adaptive to varying models,''
\newblock {\em IEEE Transactions on Pattern Analysis and Machine Intelligence},
  2021.

\bibitem{zhou2018sc2net}
Joey~Tianyi Zhou, Kai Di, Jiawei Du, Xi~Peng, Hao Yang, Sinno~Jialin Pan,
  Ivor~W Tsang, Yong Liu, Zheng Qin, and Rick Siow~Mong Goh,
\newblock ``Sc2net: Sparse lstms for sparse coding,''
\newblock in {\em Thirty-Second AAAI Conference on Artificial Intelligence},
  2018.

\bibitem{li2021learned}
Yangyang Li, Lin Kong, Fanhua Shang, Yuanyuan Liu, Hongying Liu, and Zhouchen
  Lin,
\newblock ``Learned extragradient ista with interpretable residual structures
  for sparse coding,''
\newblock in {\em Proceedings of the AAAI Conference on Artificial
  Intelligence}, 2021, vol.~35, pp. 8501--8509.

\bibitem{zheng2022hybrid}
Ziyang Zheng, Wenrui Dai, Duoduo Xue, Chenglin Li, Junni Zou, and Hongkai
  Xiong,
\newblock ``Hybrid ista: unfolding ista with convergence guarantees using
  free-form deep neural networks,''
\newblock {\em IEEE Transactions on Pattern Analysis and Machine Intelligence},
  2022.

\bibitem{ikehata2012robust}
Satoshi Ikehata, David Wipf, Yasuyuki Matsushita, and Kiyoharu Aizawa,
\newblock ``Robust photometric stereo using sparse regression,''
\newblock in {\em 2012 IEEE Conference on Computer Vision and Pattern
  Recognition}. IEEE, 2012, pp. 318--325.

\bibitem{xin2016maximal}
Bo~Xin, Yizhou Wang, Wen Gao, David Wipf, and Baoyuan Wang,
\newblock ``Maximal sparsity with deep networks?,''
\newblock in {\em Advances in Neural Information Processing Systems}, 2016, pp.
  4340--4348.

\end{thebibliography}


\begin{thebibliography}{10}

\bibitem{ablin2019learning}
Pierre Ablin, Thomas Moreau, Mathurin Massias, and Alexandre Gramfort,
\newblock ``Learning step sizes for unfolded sparse coding,''
\newblock in {\em Advances in Neural Information Processing Systems}, 2019, pp.
  13100--13110.

\bibitem{li2021learned}
Yangyang Li, Lin Kong, Fanhua Shang, Yuanyuan Liu, Hongying Liu, and Zhouchen
  Lin,
\newblock ``Learned extragradient ista with interpretable residual structures
  for sparse coding,''
\newblock in {\em Proceedings of the AAAI Conference on Artificial
  Intelligence}, 2021, vol.~35, pp. 8501--8509.

\bibitem{zheng2022hybrid}
Ziyang Zheng, Wenrui Dai, Duoduo Xue, Chenglin Li, Junni Zou, and Hongkai
  Xiong,
\newblock ``Hybrid ista: unfolding ista with convergence guarantees using
  free-form deep neural networks,''
\newblock {\em IEEE Transactions on Pattern Analysis and Machine Intelligence},
  2022.

\bibitem{liu2019alista}
Jialin Liu, Xiaohan Chen, Zhangyang Wang, and Wotao Yin,
\newblock ``Alista: Analytic weights are as good as learned weights in lista,''
\newblock in {\em International Conference on Learning Representations (ICLR)},
  2019.

\bibitem{aberdam2021ada}
Aviad Aberdam, Alona Golts, and Michael Elad,
\newblock ``Ada-lista: Learned solvers adaptive to varying models,''
\newblock {\em IEEE Transactions on Pattern Analysis and Machine Intelligence},
  2021.

\bibitem{zhang2018ista}
Jian Zhang and Bernard Ghanem,
\newblock ``Ista-net: Interpretable optimization-inspired deep network for
  image compressive sensing,''
\newblock in {\em Proceedings of the IEEE conference on computer vision and
  pattern recognition}, 2018, pp. 1828--1837.

\bibitem{liu2016robust}
Ding Liu, Zhaowen Wang, Bihan Wen, Jianchao Yang, Wei Han, and Thomas~S Huang,
\newblock ``Robust single image super-resolution via deep networks with sparse
  prior,''
\newblock {\em IEEE Transactions on Image Processing}, vol. 25, no. 7, pp.
  3194--3207, 2016.

\bibitem{chen2018theoretical}
Xiaohan Chen, Jialin Liu, Zhangyang Wang, and Wotao Yin,
\newblock ``Theoretical linear convergence of unfolded ista and its practical
  weights and thresholds,''
\newblock in {\em Advances in Neural Information Processing Systems}, 2018, pp.
  9061--9071.

\bibitem{Wu2020gated}
Kailun Wu, Yiwen Guo, Ziang Li, and Changshui Zhang,
\newblock ``Sparse coding with gated learned ista,''
\newblock in {\em Proceedings of the International Conference on Learning
  Representations}, 2020.

\bibitem{kingma2014adam}
Diederik~P Kingma and Jimmy Ba,
\newblock ``Adam: A method for stochastic optimization,''
\newblock {\em arXiv preprint arXiv:1412.6980}, 2014.

\bibitem{abadi2016tensorflow}
Mart{\'\i}n Abadi, Paul Barham, Jianmin Chen, Zhifeng Chen, Andy Davis, Jeffrey
  Dean, Matthieu Devin, Sanjay Ghemawat, Geoffrey Irving, Michael Isard,
  et~al.,
\newblock ``Tensorflow: a system for large-scale machine learning.,''
\newblock in {\em Osdi}. Savannah, GA, USA, 2016, vol.~16, pp. 265--283.

\bibitem{wu2010robust}
Lun Wu, Arvind Ganesh, Boxin Shi, Yasuyuki Matsushita, Yongtian Wang, and
  Yi~Ma,
\newblock ``Robust photometric stereo via low-rank matrix completion and
  recovery,''
\newblock in {\em Asian Conference on Computer Vision}. Springer, 2010, pp.
  703--717.

\bibitem{ikehata2012robust}
Satoshi Ikehata, David Wipf, Yasuyuki Matsushita, and Kiyoharu Aizawa,
\newblock ``Robust photometric stereo using sparse regression,''
\newblock in {\em 2012 IEEE Conference on Computer Vision and Pattern
  Recognition}. IEEE, 2012, pp. 318--325.

\end{thebibliography}

\end{document}


%
\onecolumn
\maketitle

\begin{figure*}[ht]
	\centering
	\vskip 0.0in
	\hskip 0.1in
	\subfloat[LISTA]{\label{fig:LISTA-CP}	
		\includegraphics[width=0.44\linewidth]{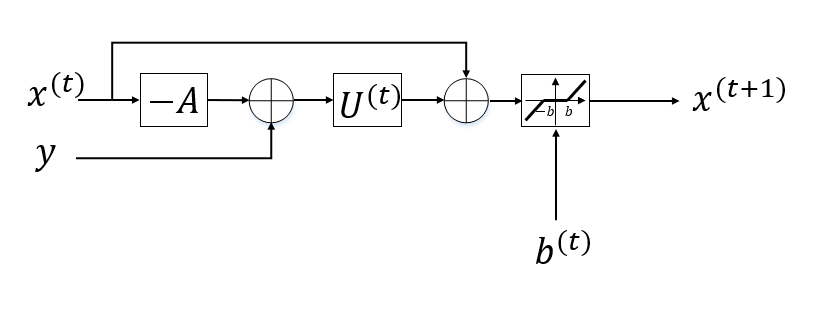}}
	\hskip 0.1in
	\subfloat[EBT-LISTA]{\label{fig:LISTA-EBT}	
		\includegraphics[width=0.43\linewidth]{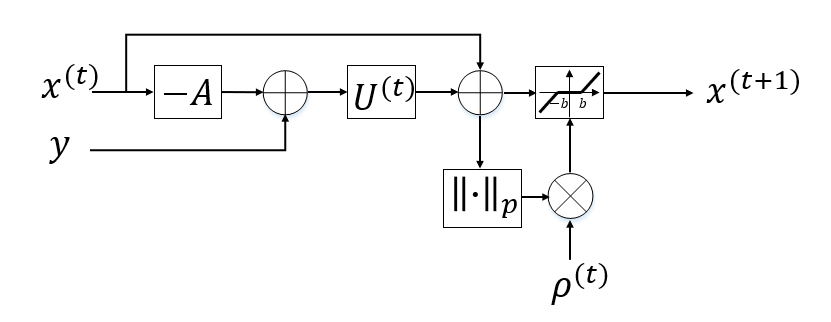}}
	\caption{The $t$-th layer of LISTA and EBT-LISTA.}\label{fig:archi}
	\vskip -0.1in
\end{figure*}
\section{Related works}
Some prior work also aims at improving LISTA.
For instance, \cite{ablin2019learning} tried only to learn the step size in ISTA, while all other parameters in the network are not learned. 
\cite{li2021learned} introduced the extragradient method in LISTA and proposed ELISTA, in which $x^{(t+\frac{1}{2})}$ is calculated to consider the curvature information. It designs multistage-thresholding functions to obtain effective sparse representation.
\cite{zheng2022hybrid} proposed hybrid ISTA and hybrid LISTA that incorporate free-form DNNs into ISTA and LISTA to improve the efficiency and flexibility without compromising the convergence rate. 
The most related work is ALISTA~\cite{liu2019alista}, in which analytic weights were obtained in a data-free manner. 
Benefiting from its training simplicity, ALISTA can be adopted to a scenario where robustness and adaptivity is required. 
We will show with experiments that our EBT can be combined with these methods to achieve further improvements.

\cite{aberdam2021ada} proposed Ada-LISTA to improve adaptivity to dictionary permutations and perturbations, which is quite different from our goal which is the adaptivity to possible variations in data (i.e., $x^\star$). 
Some other work introduced structured convolutional neural network to learn in image domains. For example, \cite{zhang2018ista} proposed ISTA-net for image compressive sensing, and \cite{liu2016robust} proposed SCN for image super-resolution. These tasks are not the main interest of this paper.

\section{Experiments}

\subsection{Basic settings}
The network architectures and training strategies in our experiments mostly follow those of prior works~\cite{chen2018theoretical,Wu2020gated}. To be more specific, all the compared networks have $d=16$ layers and the learnable parameters $W^{(t)},U^{(t)},\rho^{(t)}$ (or $b^{(t)}$ without EBT) are not shared among layers. The training batch size is 64, and we use the popular Adam optimizer~\cite{kingma2014adam} for training with its default hyper-parameters $\beta_1=0.9$ and $\beta_2=0.999$. Training is performed from layer to layer in a progressive way, i.e., if the validation loss of the current layer does not decrease for 4000 iterations, the next layer will then be trained. When training each layer, the learning rate is first initialized to 0.0005. It will first be decreased to 0.0001 and finally to 0.00001 if the validation loss does not decrease for 4000 iterations. Specifically, in the proposed methods, we impose the constraint between $W^{(t)}$ and $U^{(t)}$ and make sure it holds that $W^{(t)}=I-U^{(t)}A,\forall t$, i.e., the coupled constraints are introduced~\cite{chen2018theoretical}. All experiments are performed on NVIDIA GeForce RTX 2080 Ti using TensorFlow~\cite{abadi2016tensorflow}.

In simulation experiments, we set $m=250$, $n=500$, and we generate the dictionary matrix $A$ by using the standard Gaussian distribution. The indices of the non-zero entries in $x^\star$ are determined by a Bernoulli distribution letting its sparsity (i.e., the probability of any of its entries be zero) be $p_b$, while the magnitudes of the non-zero entries are sampled from the standard Gaussian distribution. The noise $\varepsilon$ is sampled from a Gaussian distribution where the standard deviation is determined by the noise level. With $y=Ax^\star+\varepsilon$, we can randomly synthesize in-stream $x^\star$ and get a corresponding set of observations $y$ for training, thus the number of training samples grows as the training proceeds. We also synthesize two sets for validation and test, respectively, each containing 1000 samples. The sparse coding performance of different models is evaluated by the normalized mean squared error (NMSE) in decibels (dB):
\begin{equation}
\label{NMSE}
\textup{NMSE}(x,x^\star)=10\log_{10}\left(\frac{{\left\|x-x^\star\right\|}_2^2}{{\left\|x^\star\right\|}_2^2}\right).
\end{equation}

\subsection{Simulation Results for Showing Disentanglement}
\begin{figure*}[ht]
	\centering
	\subfloat[LISTA]{\label{fig:valid3}	
		\includegraphics[width=0.42\linewidth]{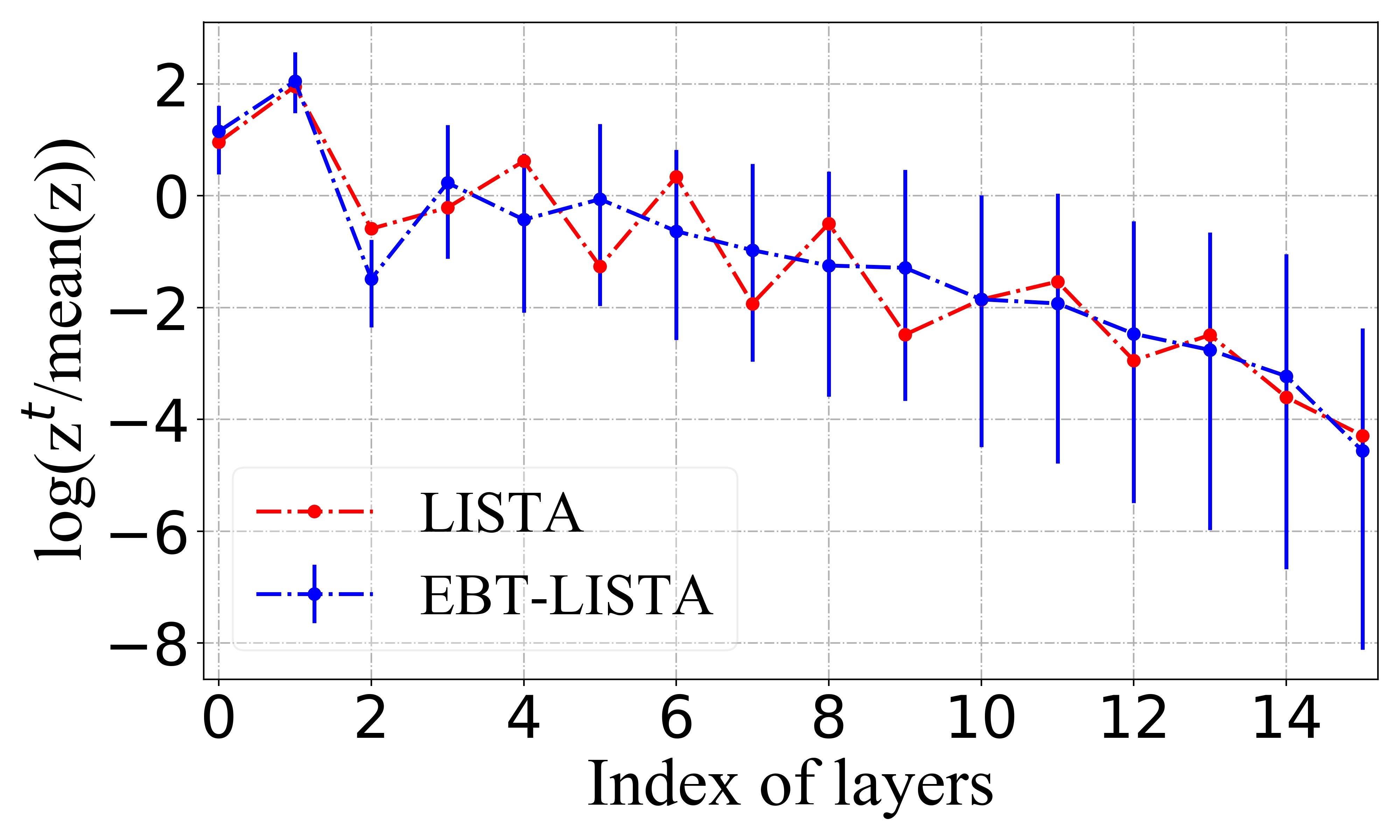}}
	\hskip 0.2in
	\subfloat[LISTA-SS]{\label{fig:valid4}	
		\includegraphics[width=0.42\linewidth]{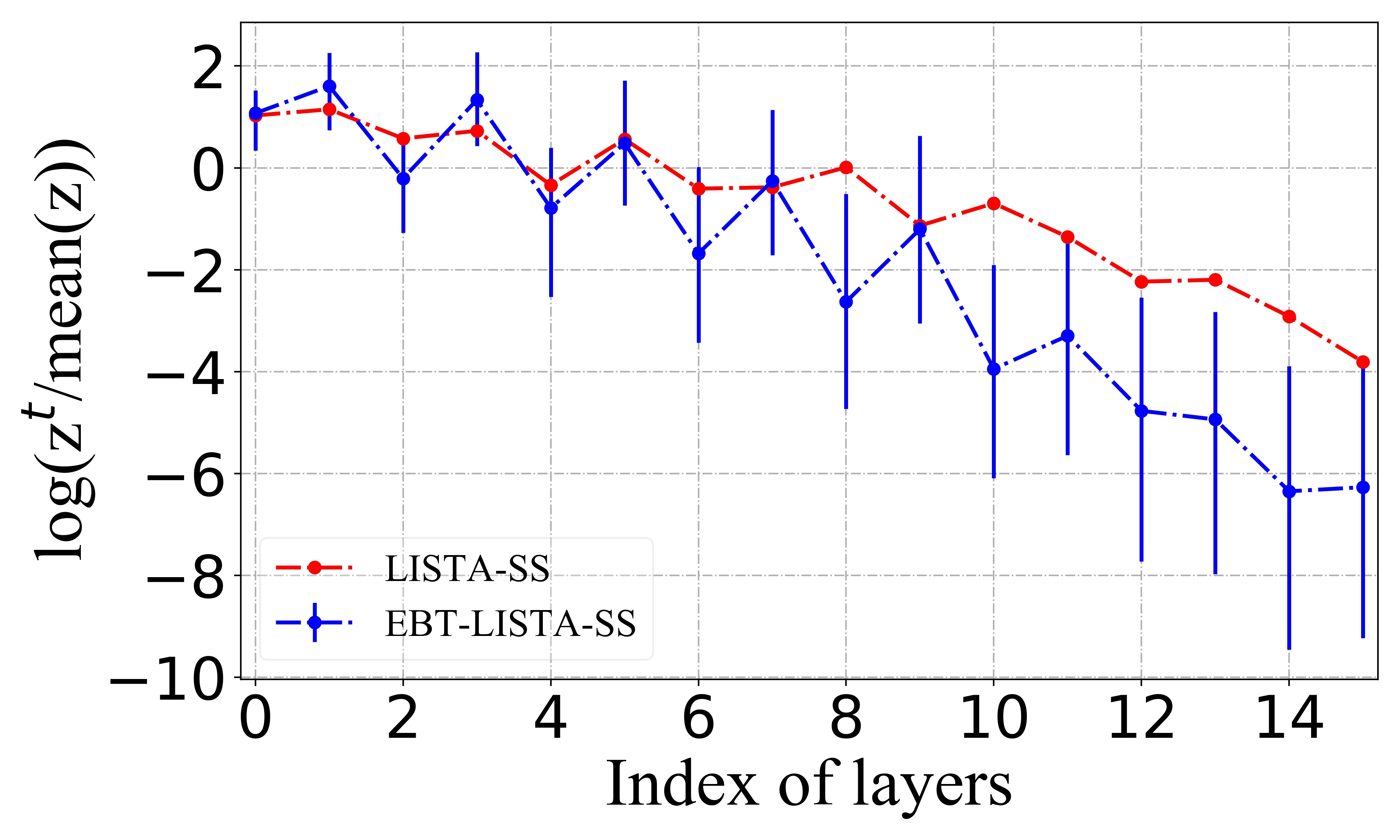}}
	\caption{Thresholds obtained from different methods across layers.}.
	\label{fig:thresholds}
\end{figure*}

We analyze the obtained threshold values in EBT-LISTA and EBT-LISTA-SS, i.e., $b^{(t)}=\rho^{(t)}\|U^{(t)}(Ax^{(t)}-y)\|_\phi$ (with $\phi=2$), and compare them with the thresholds values obtained in LISTA and LISTA-SS. Note that the threshold values in our EBT-based models differ from sample to sample, we show the results in Figure~\ref{fig:thresholds}. It can be seen that the learned thresholds in our EBT-based methods and the original LISTA and LISTA-SS are similar, which indicates that the introduced EBT mechanism does not modify the training dynamics of the original methods, and our EBT works by disentangling the reconstruction error and learnable parameters.

\subsection{Validation of Theorem \ref{thm2}}
\begin{figure}[ht]
	\centering
	\vskip -0.15in
	\subfloat[LISTA-SS]{\label{fig:LISTA_fun}	
		\includegraphics[width=0.4\linewidth]{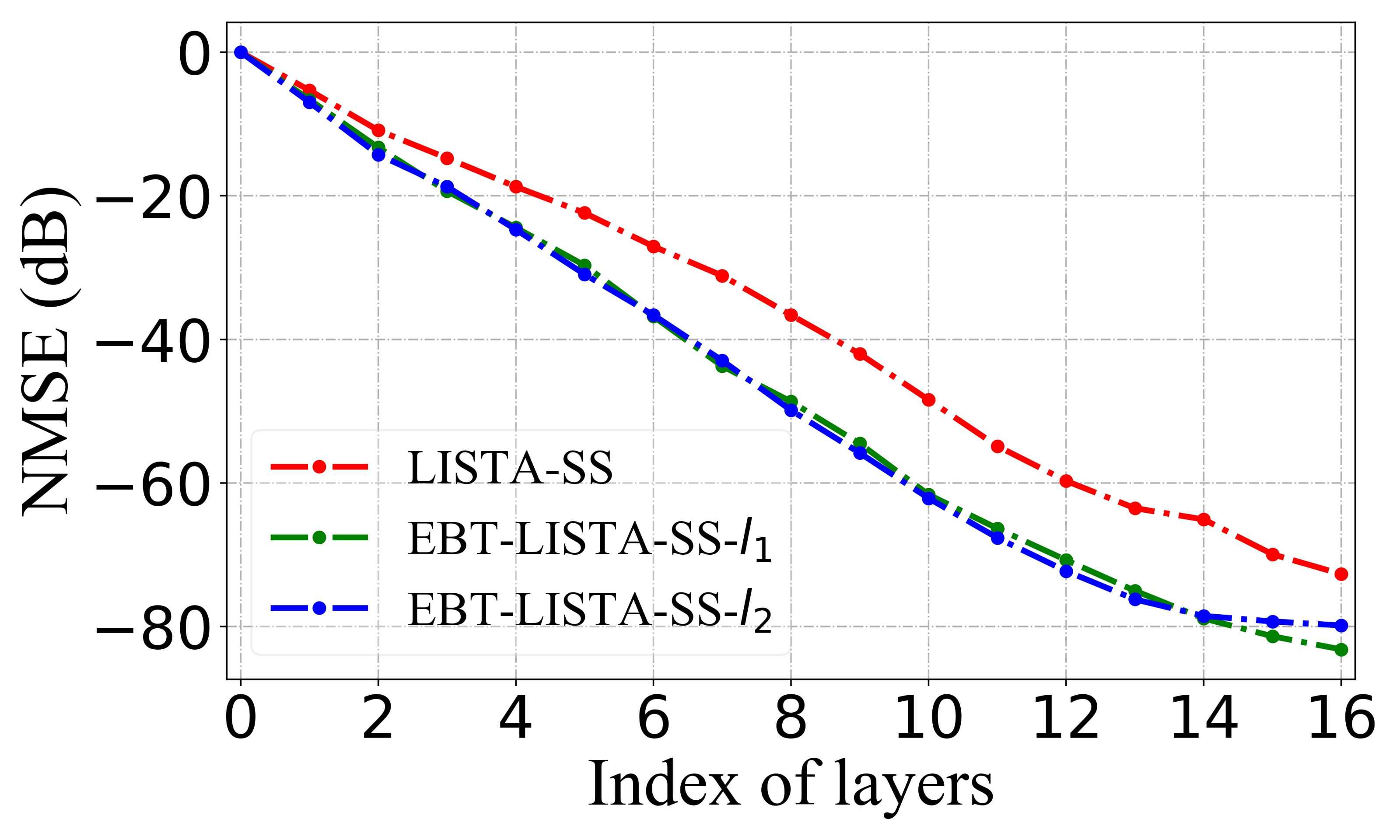}}
	\hskip 0.1in
	\subfloat[ALISTA]{\label{fig:ALISTA_fun}
		\includegraphics[width=0.40\linewidth]{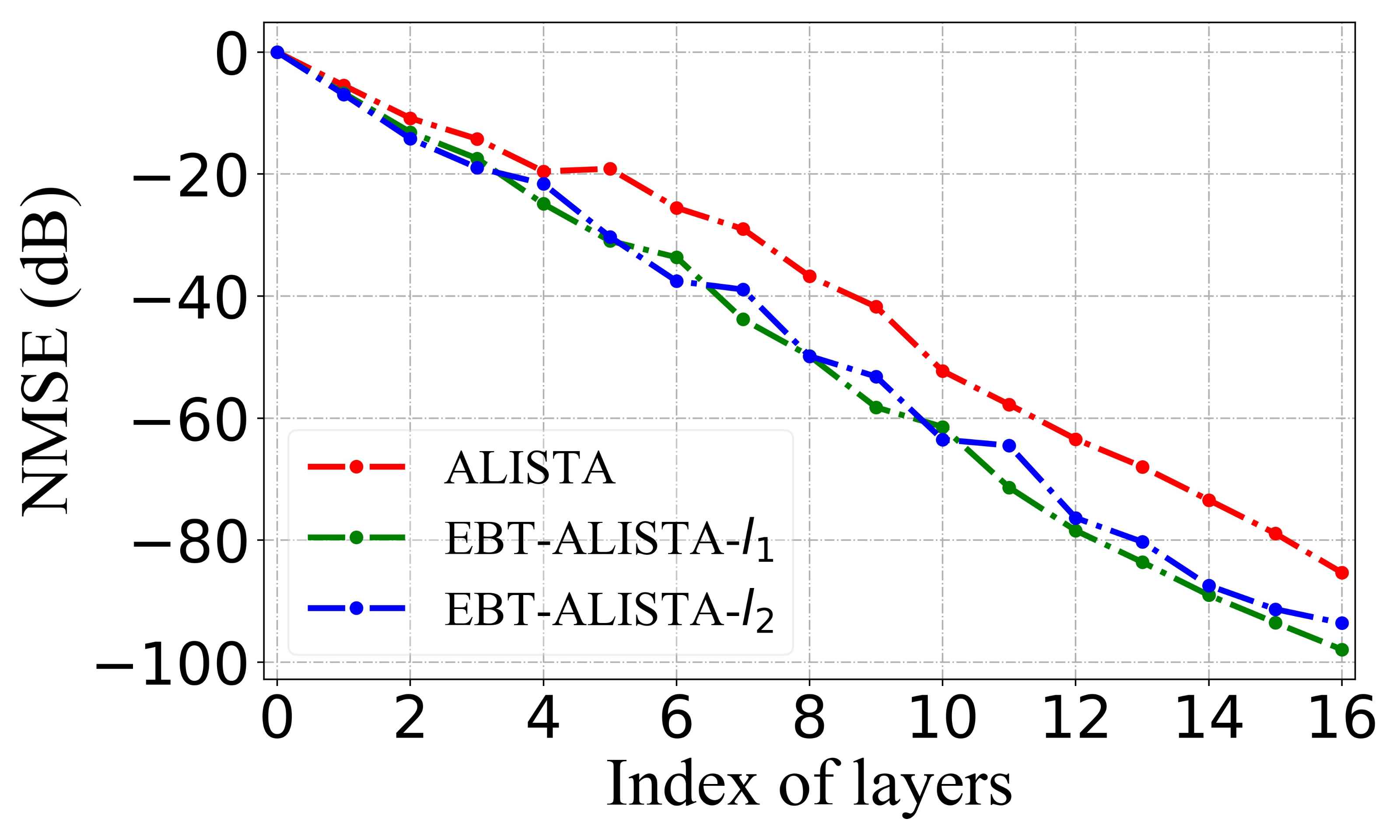}}	
	\caption{Validation of Theorem \ref{thm2}: there exist two convergence phases and our EBT accelerates the convergence of LISTA-SS, in particular in the first phase.} 
	\label{fig:fun} \vskip -0.0in
\end{figure}

Figure~\ref{fig:LISTA_fun} shows how the NMSE of EBT-LISTA-SS varies across layers. In addition to the $l_1$ norm (i.e., $b^{(t)}=\rho^{(t)}\|U^{(t)}(Ax^{(t)}-y)\|_1$) concerned in the theorem, we also test EBT-LISTA-SS with the $l_2$ norm (i.e., by letting $b^{(t)}=\rho^{(t)}\|U^{(t)}(Ax^{(t)}-y)\|_2$). It can be seen that, with both the $l_1$ and $l_2$ norms, EBT-LISTA-SS leads to consistently faster convergence than LISTA-SS. Also, it is clear that there exist two convergence phases for EBT-LISTA-SS and LISTA-SS, and the later phase is indeed faster than the earlier phase. With faster convergence, EBT-LISTA-SS finally achieves superior performance.
The experiment is performed in the noiseless case with $p_b=0.95$. 
Similar observations can be made on other variants of ALISTA (e.g., ALISTA~\cite{liu2019alista}, see Figure~\ref{fig:ALISTA_fun}).

\begin{figure}[ht]
	\centering
	\vskip -0.15in
	\subfloat[Uniform distribution]{\label{fig:uniform}
		\includegraphics[width=0.4\linewidth]{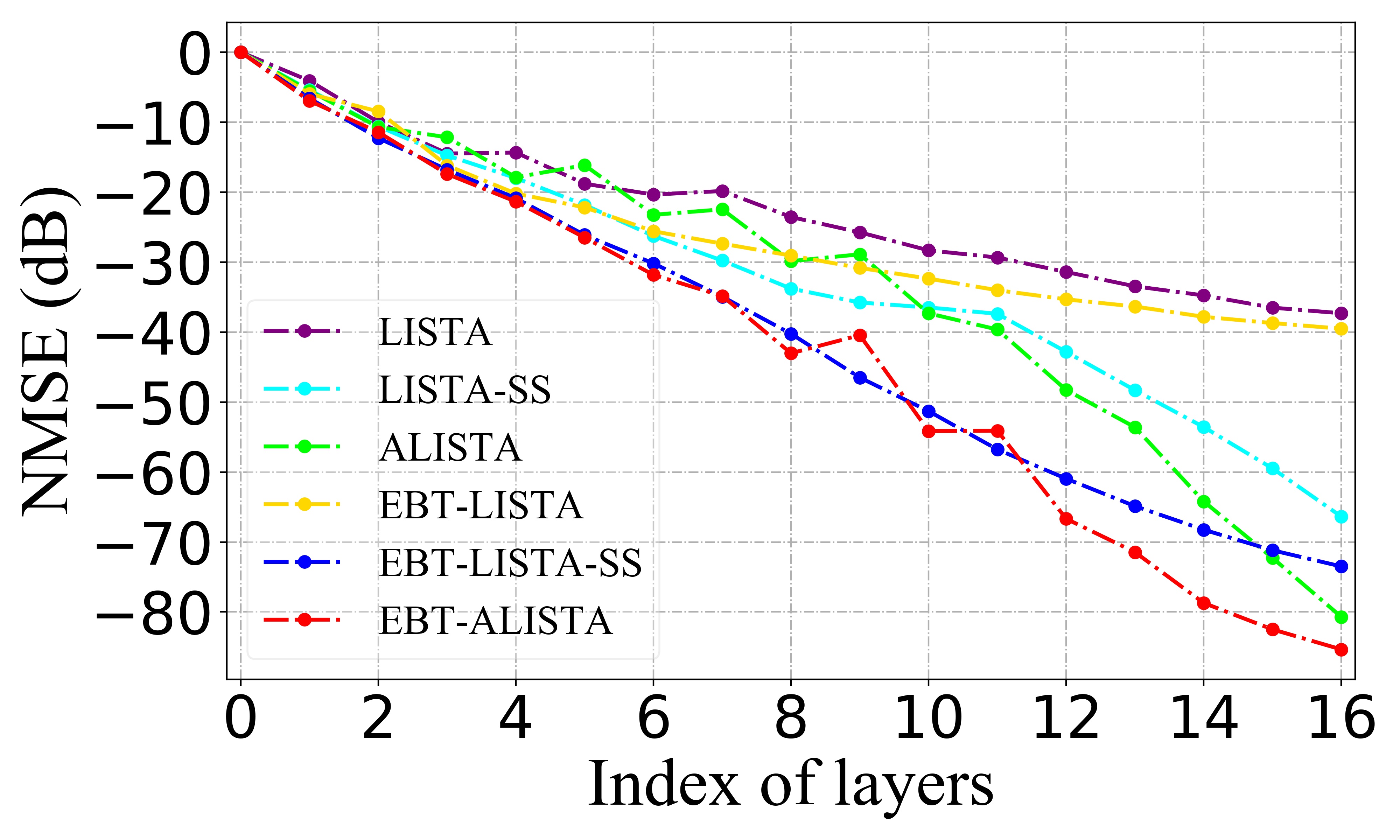}}
	\subfloat[Truncated Gaussian distribution]{\label{fig:normal}
		\includegraphics[width=0.4\linewidth]{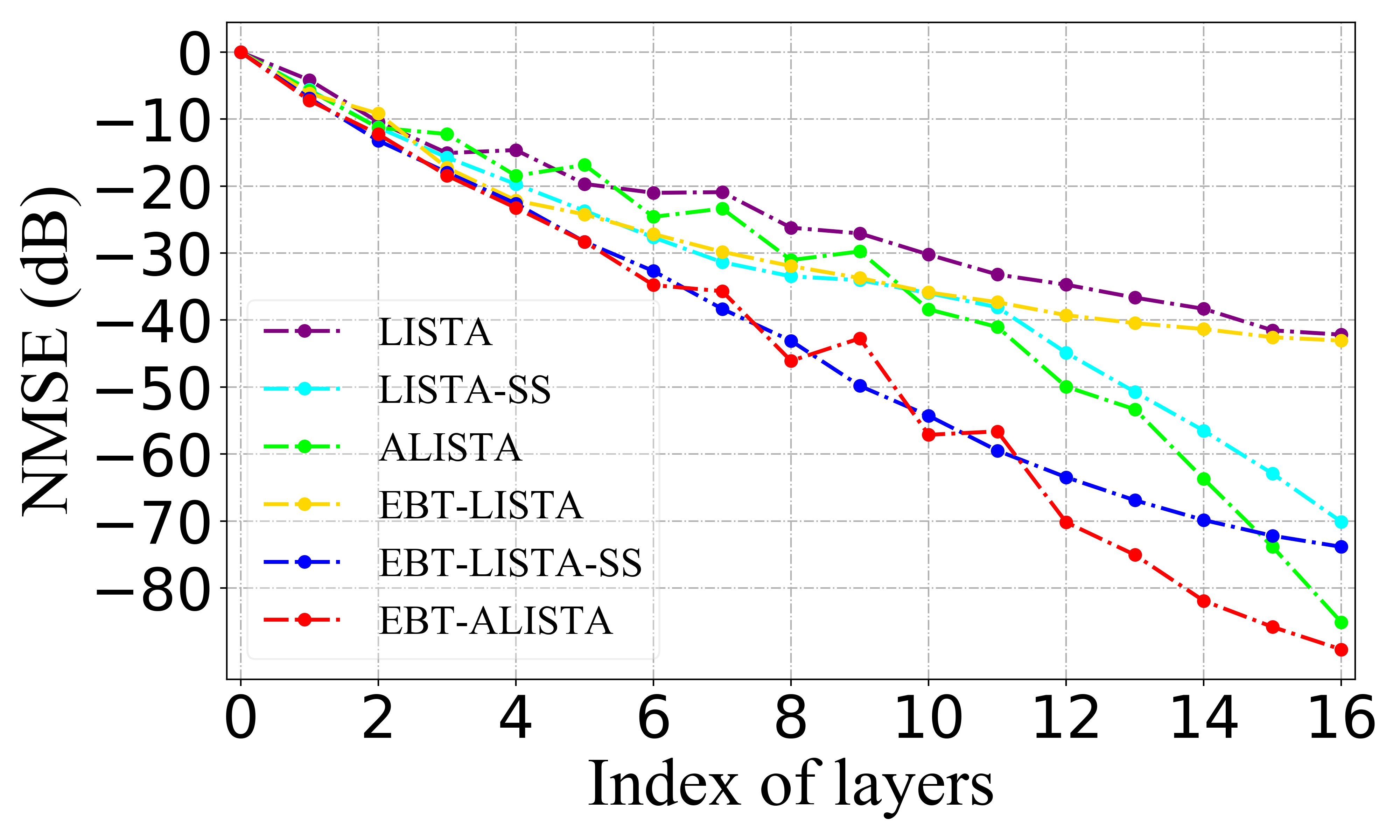}}
	\caption{NMSE of different sparse coding methods when the sparsity of the data follows a certain distribution.}
	\vskip -0.2in
	\label{fig:T1}
\end{figure}

\subsection{Random sparsity}\label{sec:randomSparsity}
We here consider the scenario where the sparsity of data follows a certain distribution. We test with two distributions of sparsity (i.e., $p_b$): $p_b\sim U(0.9,1)$ (uniform distribution) and $p_b\sim N(0.95,0.025)$ with an constraint of $p\in[0.9,1]$ (truncated Gaussian distribution). The comparison results in such settings are shown in Figure \ref{fig:T1}. Our EBT leads to huge advantages in all the models and settings, indicating that the conclusion in Theorem 1 can be extended to broader distributions of data sparsity.

\subsection{Additional Comparison with competitors} 
\begin{figure*}[ht]
    \centering
    \centering
	\subfloat[SNR=$\infty$]{\label{fig:inf1}	
		\includegraphics[width=0.30\linewidth]{pic/inf.jpg}}
	\hskip 0.1in
	\subfloat[SNR=40dB]{\label{fig:40}
	    \includegraphics[width=0.31\linewidth]{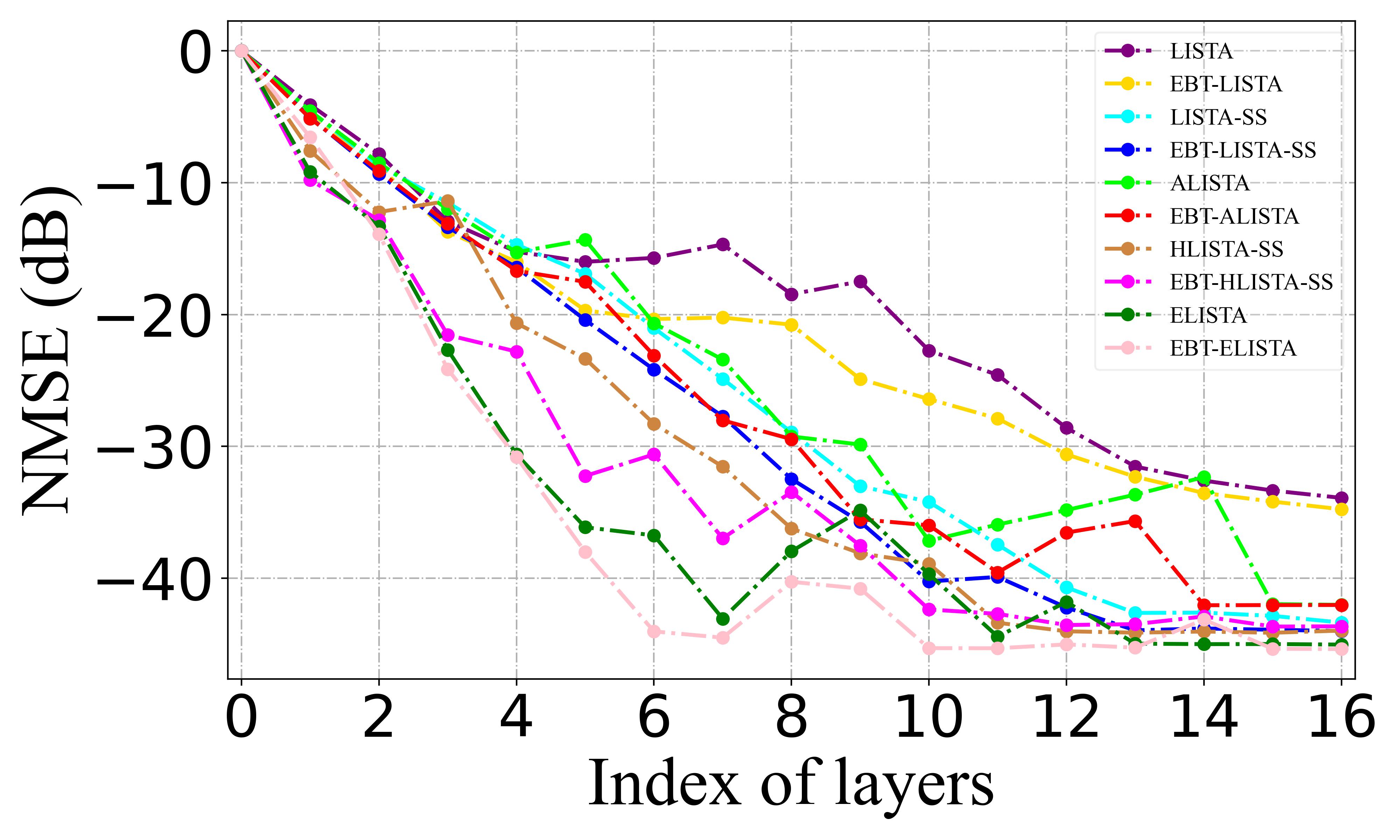}}	
	\hskip 0.1in
	\subfloat[SNR=20dB]{\label{fig:20}
		\includegraphics[width=0.31\linewidth]{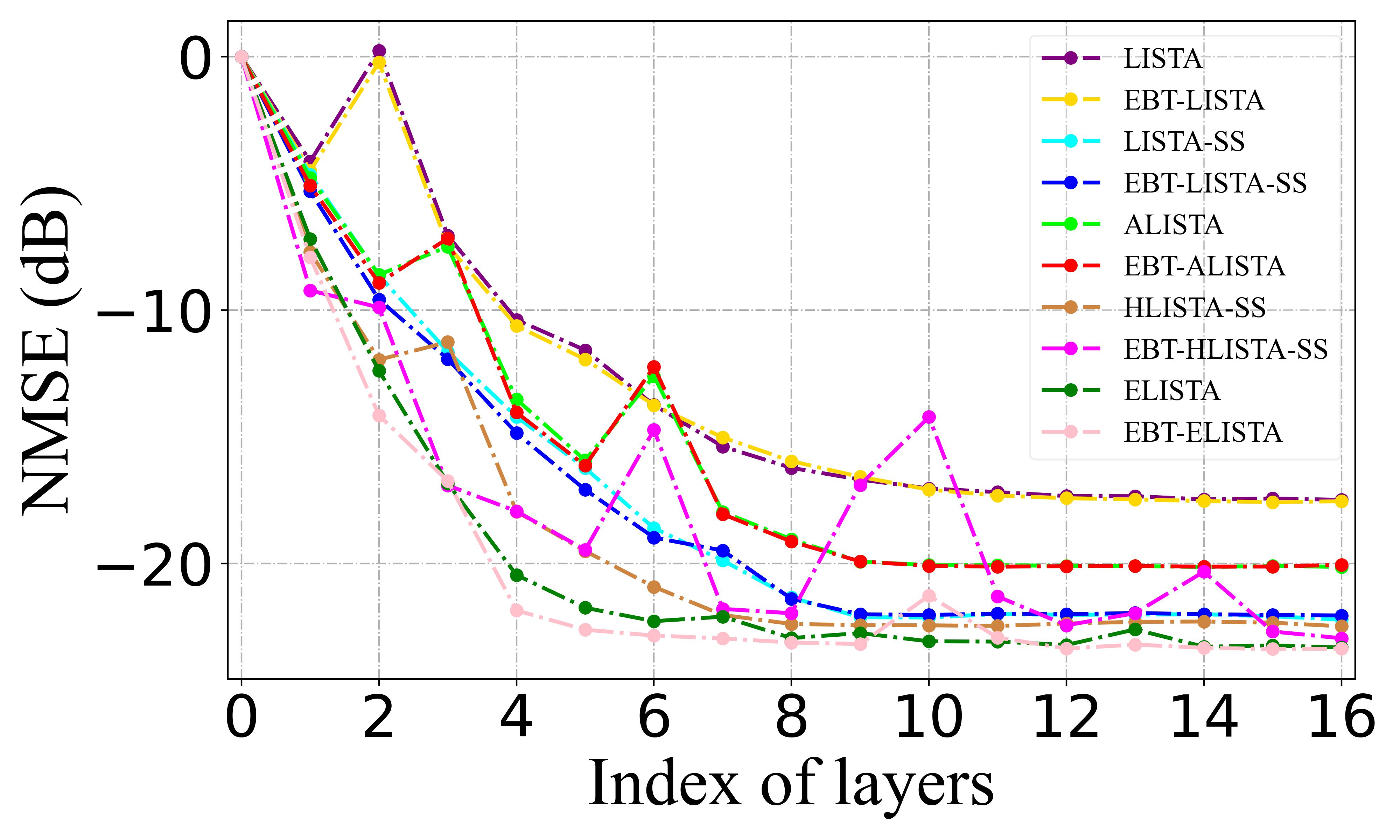}}
	\caption{NMSE of different sparse coding methods under different noise levels with $p_b=0.9$. It can be seen that our EBT performs favorably well under SNR=$\infty$ and SNR=40dB.}
	\label{fig:noise} \vskip 0.0in
\end{figure*}

\begin{figure*}[ht]
    \centering
	\subfloat[condition number=3]{\label{fig:P10}
		\includegraphics[width=0.31\linewidth]{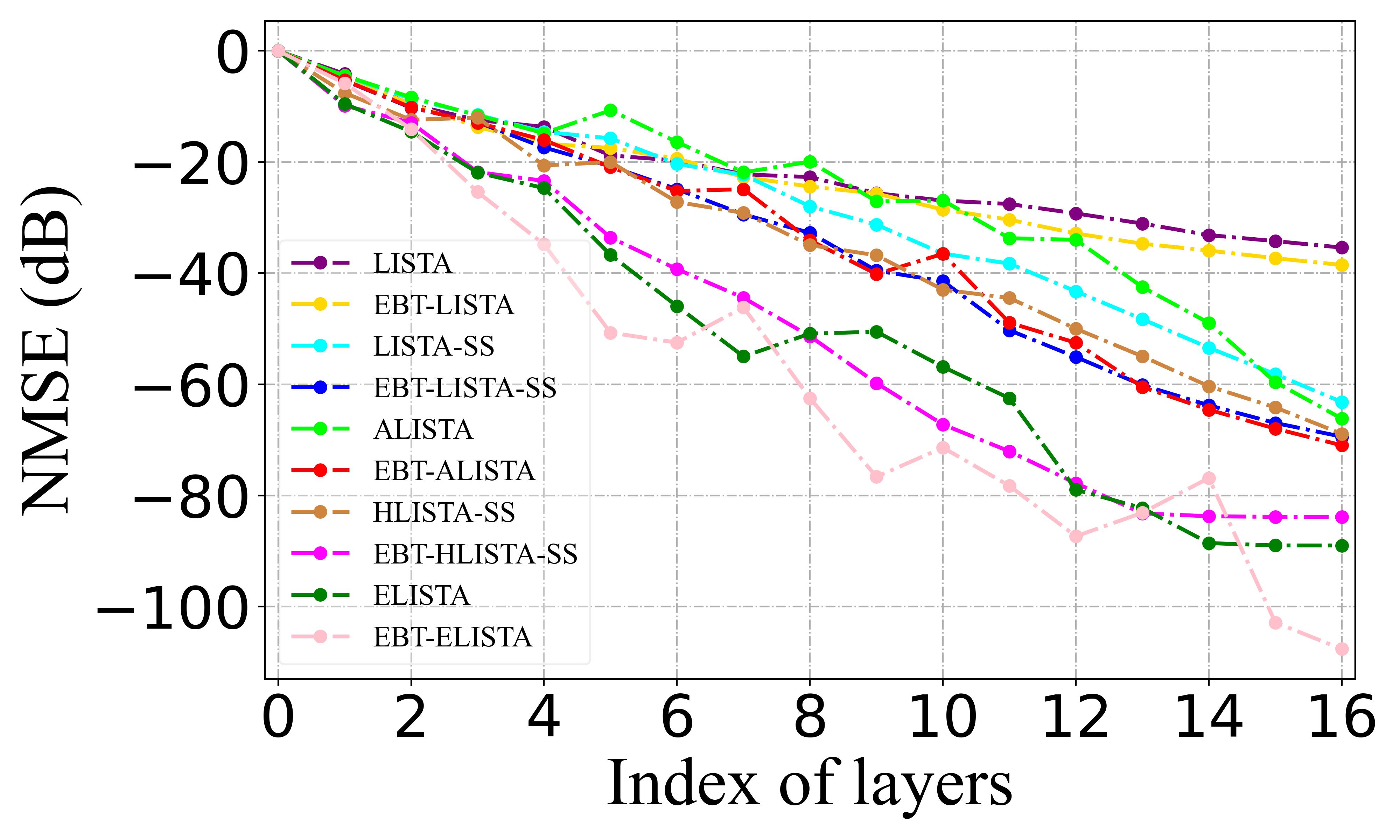}}
	\hskip 0.1in
	\subfloat[condition number=30]{\label{fig:C30}
		\includegraphics[width=0.31\linewidth]{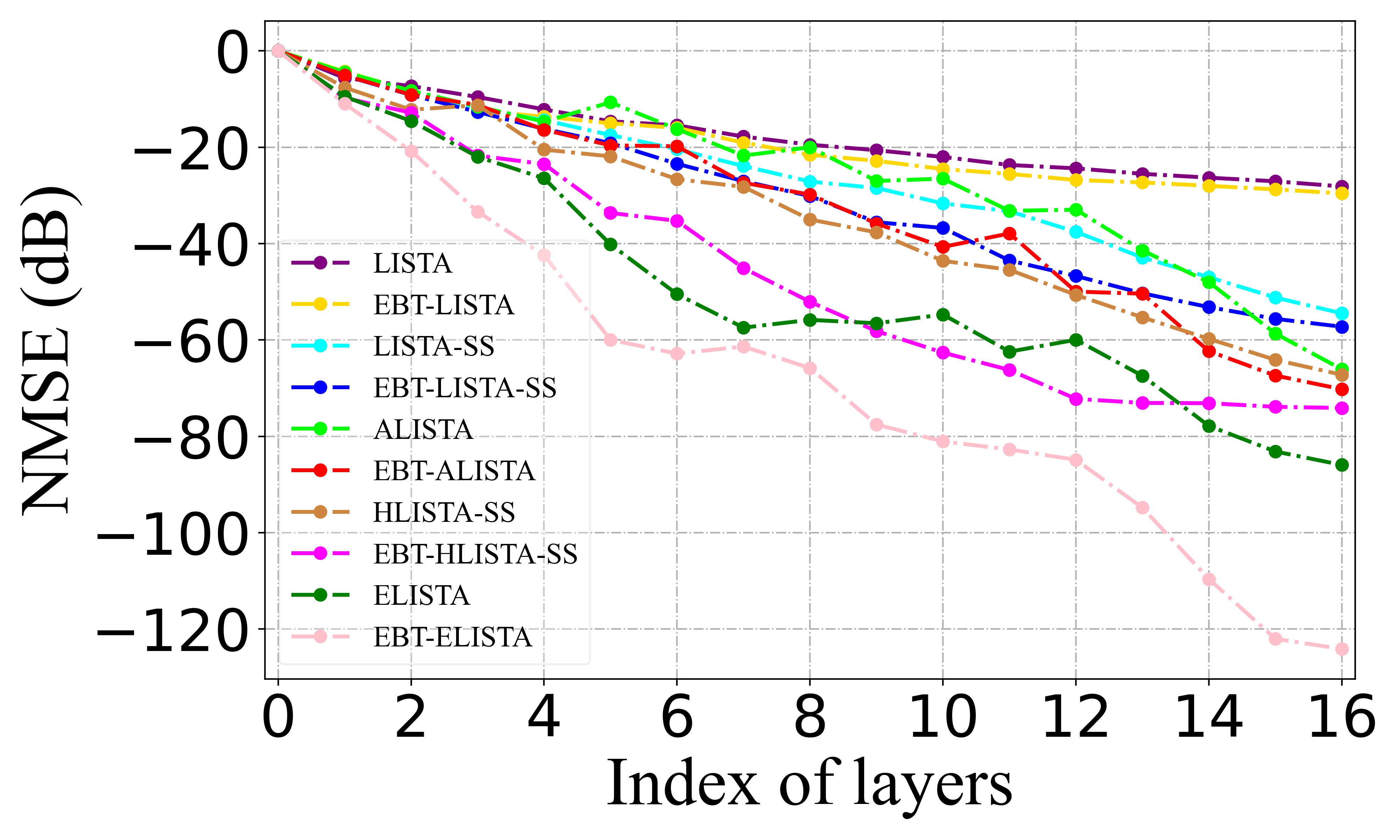}}
	\hskip 0.1in
	\subfloat[condition number=100]{\label{fig:C100}
		\includegraphics[width=0.31\linewidth]{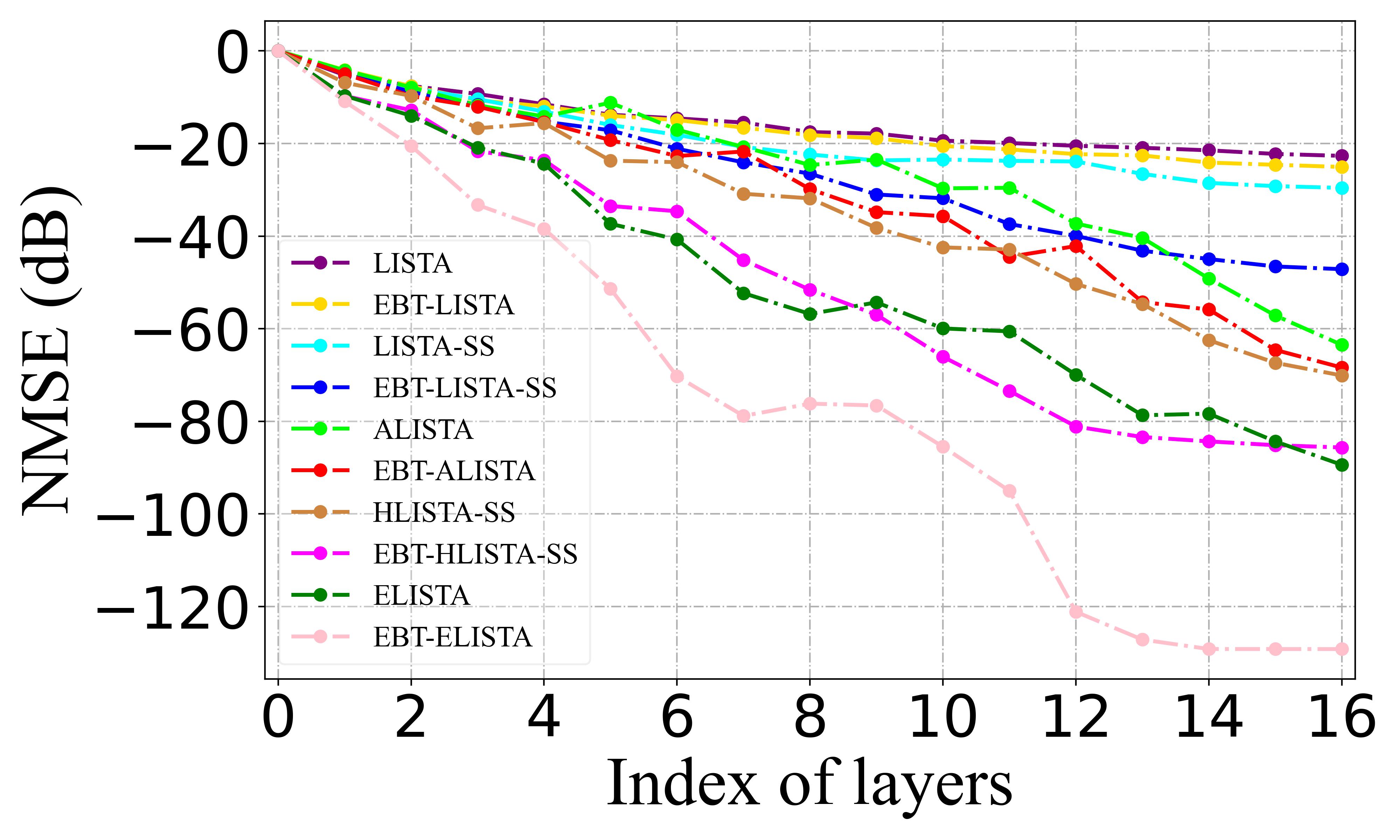}}
	\caption{NMSE of different sparse coding methods under different condition numbers with $p_b=0.9$.}
	\label{fig:sets}
\end{figure*}

More experiment results than Figure 2 in the main paper are given here. The performances of different networks under different noise levels (with $p_b=0.9$) are shown in Figure~\ref{fig:noise}. It can be seen that when combined with LISTA and its variants, our EBT achieves better or similar performance.
The figures show that the performance of our EBT is more promising in the noiseless or low noise cases (i.e., SNR=$\infty$ and SNR=40dB), while in very noisy scenarios it provides little help.
Figure \ref{fig:sets} shows the results of our methods with EBT (i.e. EBT-LISTA, EBT-LISTA-SS, EBT-ALISTA, EBT-ELISTA, and EBT-HLISTA-SS) and other competitors under different condition number (with $p_b=0.9$). From Figure \ref{fig:sets}, we can find that our EBT leads to better performance as shown in the results in the main paper.

\begin{figure*}[t]
	\centering
	\subfloat[$p_e$=0.8, LISTA-SS: $\zeta$=0.055]{\label{fig:p02}
		\includegraphics[width=0.44\linewidth]{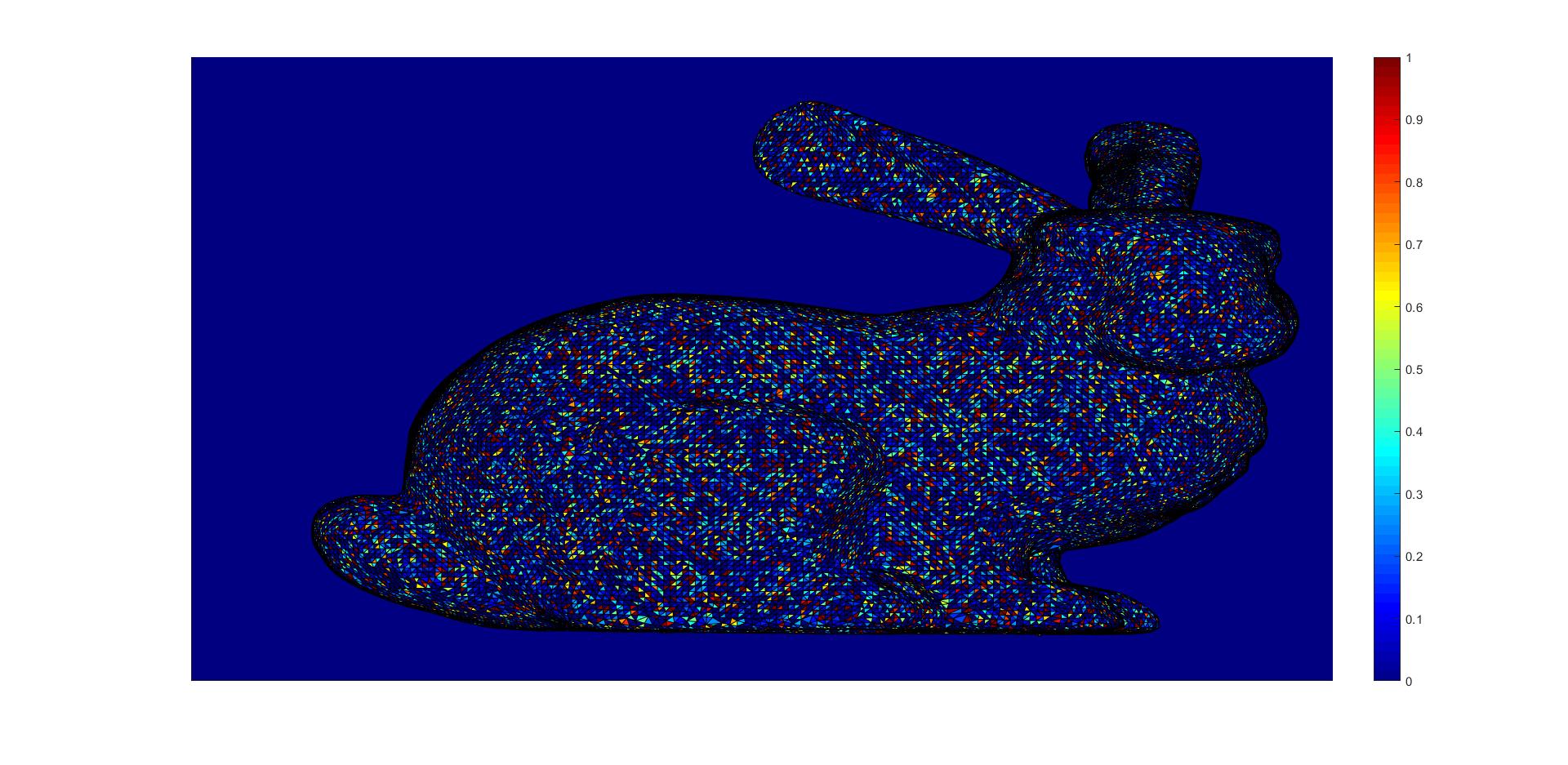}}
	\hskip 0.0in
	\subfloat[$p_e$=0.8, EBT-LISTA-SS: $\zeta$=0.041]{\label{fig:p02ebt}
		\includegraphics[width=0.44\linewidth]{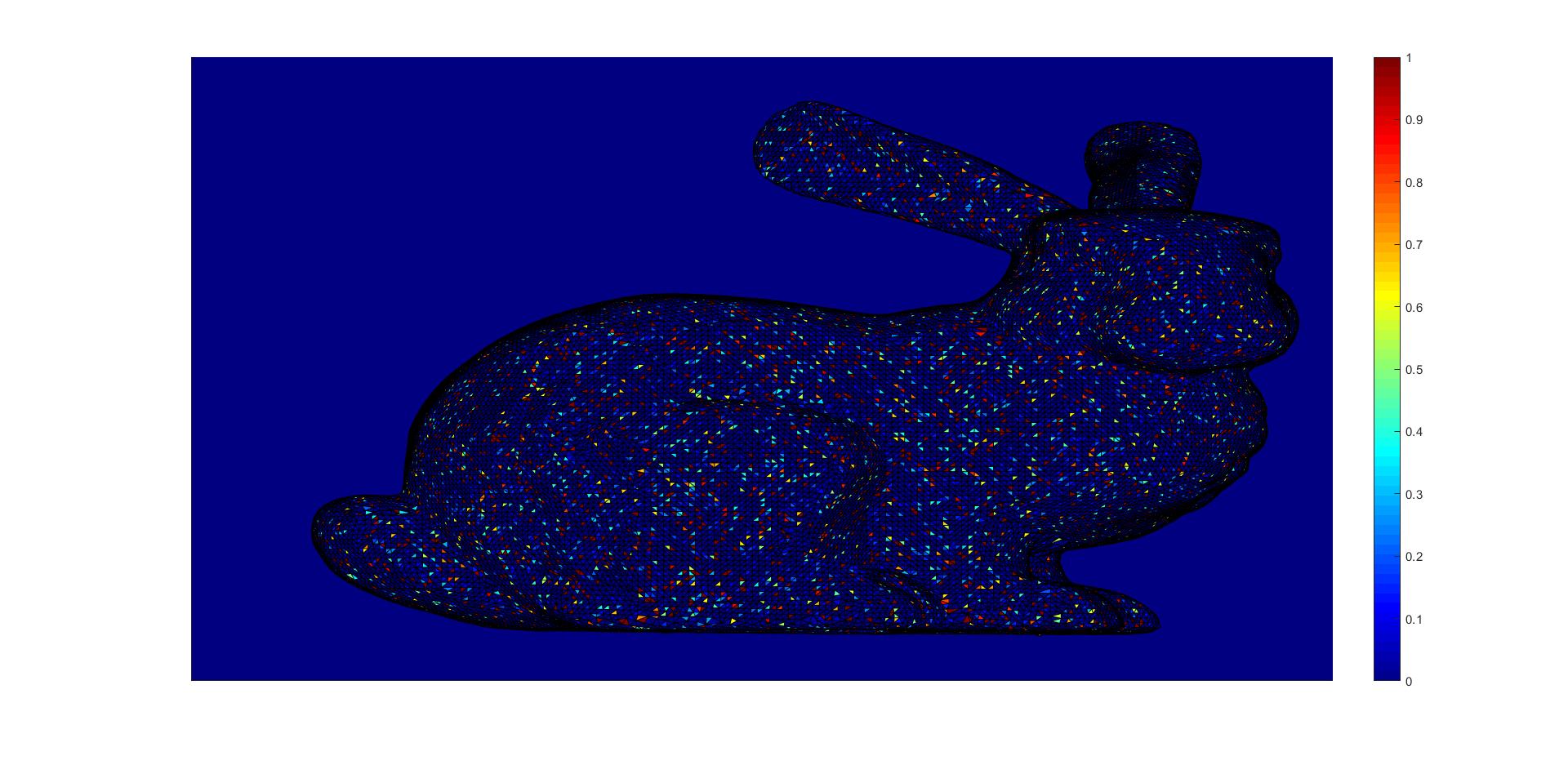}}
	\vskip 0.05in
	\subfloat[$p_e$=0.9, LISTA-SS: $\zeta$=0.0067]{\label{fig:p01}
		\includegraphics[width=0.44\linewidth]{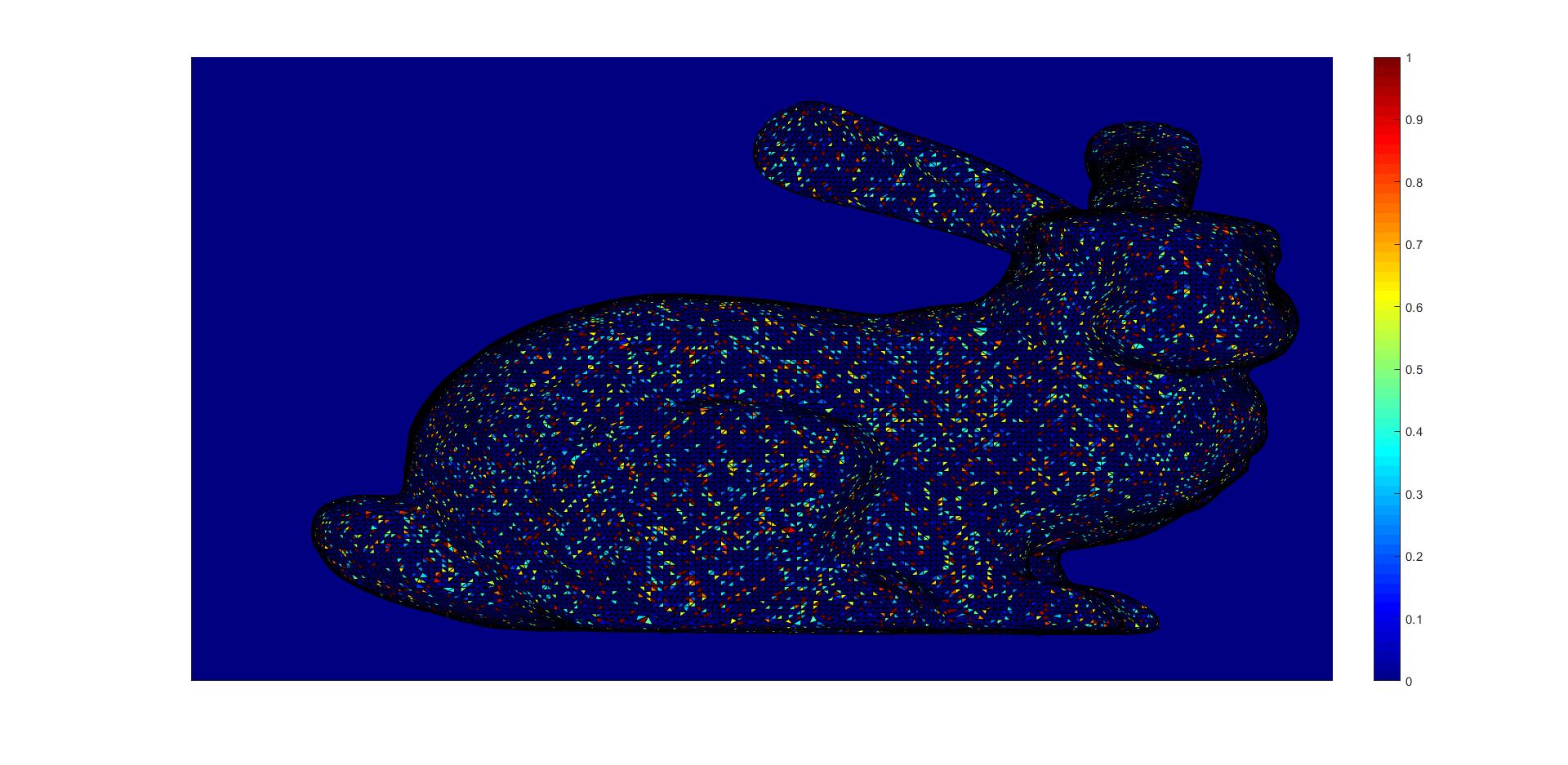}}
	\hskip 0.0in
	\subfloat[$p_e$=0.9, EBT-LISTA-SS: $\zeta$=0.0026]{\label{fig:p01ebt}
		\includegraphics[width=0.44\linewidth]{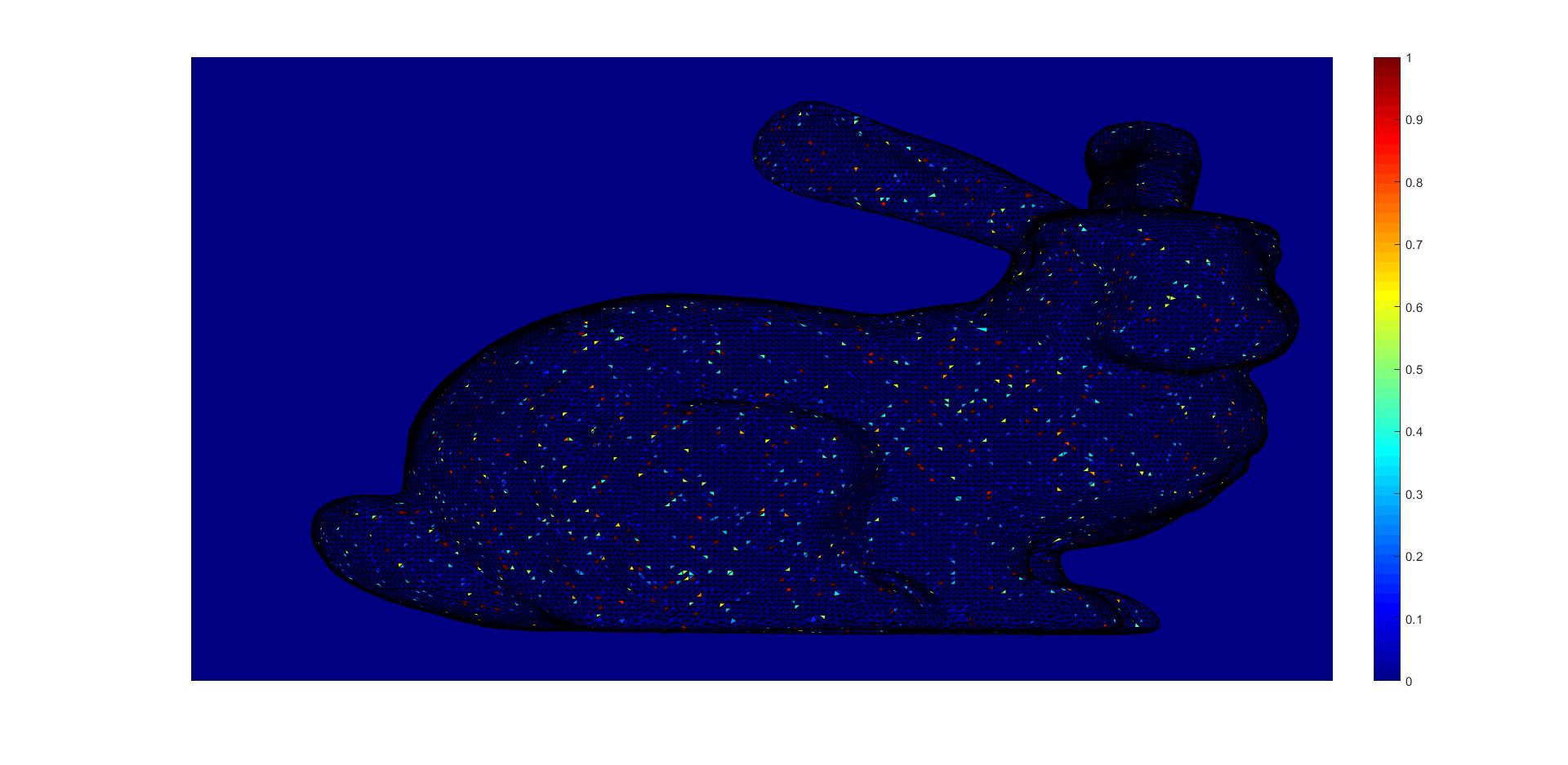}}
	\vskip -0.05in
	\caption{Reconstruction 3D error maps of different methods in different settings. $\zeta$ here is the mean estimation error in degree. Note that the maximal error is 0.1 and 0.03 in theory when $p_e=0.8$ and $p_e=0.9$, respectively. It can be seen that with EBT-LISTA-SS considerably outperforms LISTA-SS in this task.} \vskip -0.1in %
	\label{fig:bunny}
\end{figure*}
\subsection{Photometric Stereo Analysis}
Here we give a detailed discussion on photometric stereo analysis. To be specific, the task solves the problem of estimating the normal direction of a Lambertian surface, given $q$ observations under different light directions. It can be formulated as
\begin{equation}
\label{psa}
o = \rho Lv+e,
\end{equation}
where $o \in \mathbb R^{q}$ is the observation, $v\in\mathbb R^{3}$ represents the normal direction of the Lambertian surface to be estimated, $L\in\mathbb R^{q\times 3}$ represents the normalized light directions, $e$ is the noise vector, and $\rho$ is the diffuse albedo scalar. Although the normal vector $v$ is unconstrained in Eq.~(\ref{psa}), the noise vector $e$ is found to be generally sparse~\cite{wu2010robust,ikehata2012robust}. Therefore, we may estimate the noise $e$ first. We introduce the orthogonal complement of $L$, denoted by $L^{\dagger}$, to rewrite Eq.~(\ref{psa}) as
\begin{equation}
L^{\dagger}o = \rho L^{\dagger}Lv + L^{\dagger}e = L^{\dagger}e.
\end{equation}
On the basis of the above equation, the estimation of $e$ is basically a noiseless sparse coding problem, where $L^{\dagger}$ is the dictionary matrix $A$, $e$ is the sparse code $x^\star$ to be estimated in the reformulated problem, and $L^{\dagger}o$ is the observation $y$. Once we have gotten a reasonable estimation of $e$, we can further obtain $v$ by using the equation $v=L^{\dagger}(o-e)$.

Other than Table 1 in main paper, we here build the reconstruction 3D error maps for LISTA-SS and EBT-LISTA-SS when $p_e=0.8$ and $p_e=0.9$, as shown in Figure \ref{fig:bunny}. Note that in the picture, brighter means larger estimation error. The results from the figure also show that EBT-LISTA-SS outperforms LISTA-SS in photometric stereo analysis.

\subsection{EBT mechanism on (F)ISTA} 
\begin{figure}[ht]
    \centering
	\vskip -0.05in
	\subfloat[$\lambda$=0.1]{\label{fig:ISTA_01}
		\includegraphics[width=0.4\linewidth]{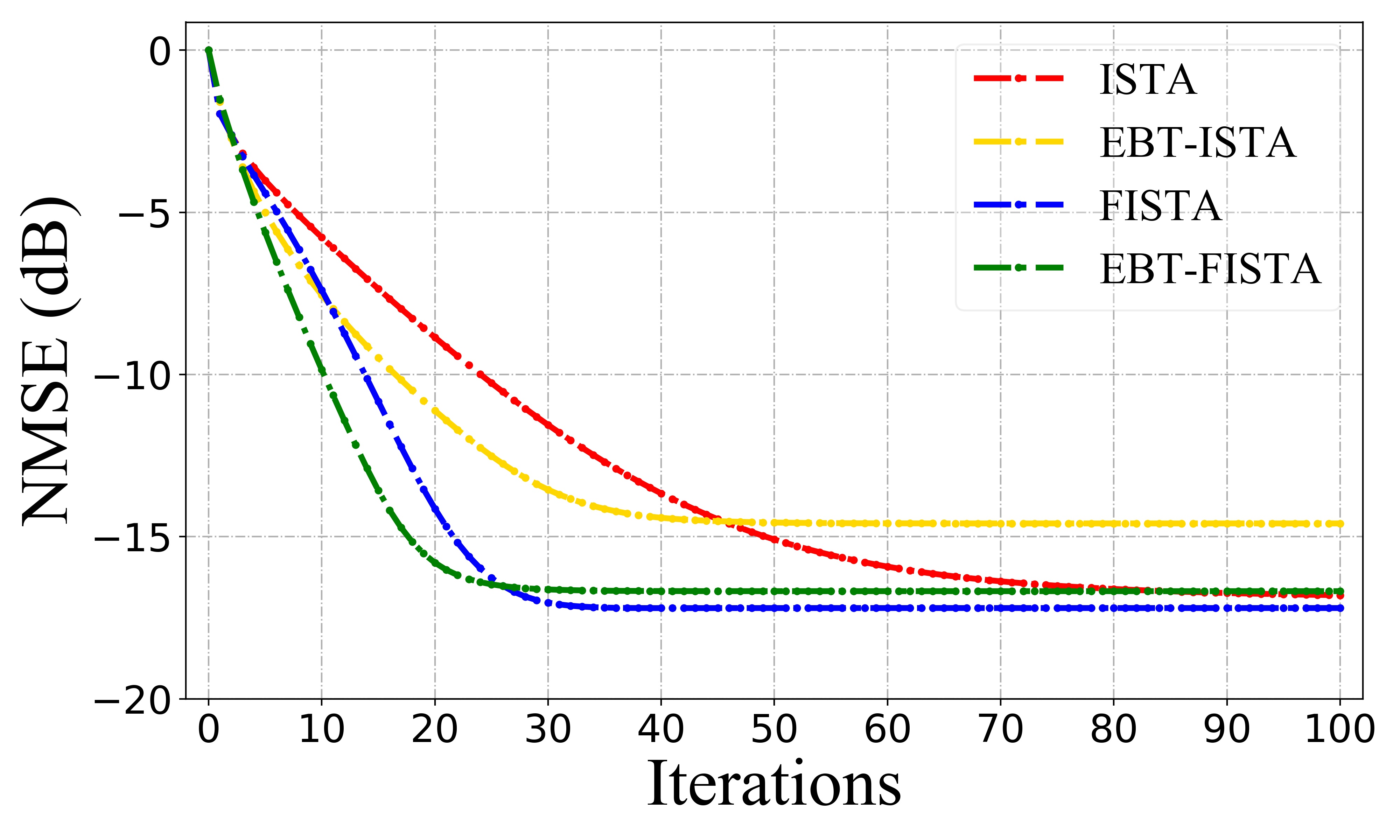}}
	\hskip 0.1in
	\subfloat[$\lambda$=0.2]{\label{fig:ISTA_02}
		\includegraphics[width=0.4\linewidth]{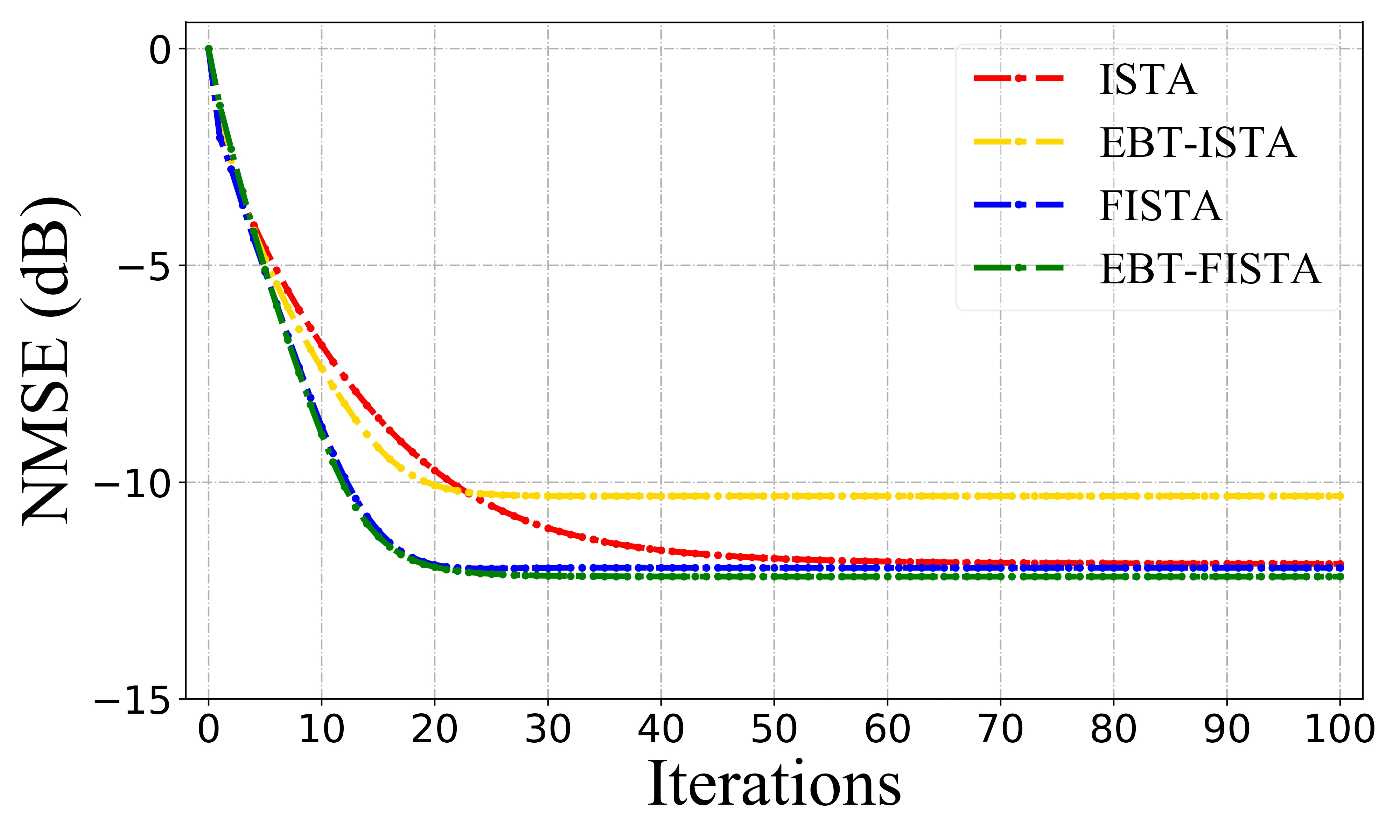}}
	\caption{NMSE of different sparse coding methods where different regularization coefficients $\lambda$ are considered.}
	\label{fig:ISTA}
\end{figure}

We further test the proposed EBT mechanism on the standard ISTA and FISTA.  We set the regularization coefficients $\lambda$ in the Lasso problem (and ISTA algorithm) as $0.1$ and $0.2$, respectively. With EBT incorporated, we use $\lambda\|Ax^{(t)}-y\|_1/\gamma$ as the threshold at the $t$-th layer, and we use $\lambda/\gamma$ for (F)ISTA. From the experiment results shown in Figure~\ref{fig:ISTA}, we can find that our EBT mechanism leads to faster convergence, however, the final performance is not satisfactory. This can possibly be ascribed to the convergence speed, considering that LISTA has linear convergence in theory while ISTA and FISTA converge in a sub-linear manner.

\section{Theoretical Analysis}
Here, we provide theoretical analyses of the theorems shown in the main paper.
Before delve deep into the proof, we first give some important notations.
We use $\mathrm{supp}(x)$ to represent the support set of vector $x$. We denote $\mathcal{S}$ as the support set of $x^\star$, and $|\mathcal{S}|$ is the number of the elements in the set $\mathcal{S}$. We use $x_i$ denotes the $i$-th element of a vector $x$, and $A_{ij}$ denotes the element of matrix $A$ placed on $i$-th raw and $j$-th column.

\subsection{Proof of Theorem 1}
Remind that the our EBT-LISTA is formulated as $x^{(0)}=0$, and, for $t=0,\dots,d$,
\begin{equation}
\label{thresh_2}
\begin{aligned}
x^{(t+1)}&=\mathrm{sh}_{b^{(t)}}((I-U^{(t)}A)x^{(t)}+U^{(t)}y),\\
b^{(t)}&=\rho^{(t)}\|U^{(t)}(Ax^{(t)}-y)\|_\phi.
\end{aligned}
\end{equation}

\begin{Theorem}
	\label{thm1}
	\rm{\textbf{(Convergence of EBT-LISTA)}}
	For EBT-LISTA formulated in Eq.~(\ref{thresh_2}) where $\phi=1$, $x^\star$ is sampled from $\gamma(B,s)$. If $s$ is small such that $\mu(A)(2s-1) < 1$, $U^{(t)} \in \mathcal{W}(A)$ and $\rho^{(t)} = \frac{\mu(A)}{1-\mu(A)s}$,
	the estimation $x^{(t)}$ at the $t$-th layer satisfies $$\|x^{(t)}-x^\star\|_2\leq q_0\exp{(c_1t)},$$ where $q_0<sB$ and $c_1<c_0$ hold with the probability of $1 - \mu(A)s$.
\end{Theorem}

\textbf{Proof.} The proving process of our Theorem 1 is similar to Theorem 2 in~\cite{chen2018theoretical}.

We first prove the no "false positive" property, i.e., $\forall t \geq 0$, we have $\mathrm{supp}(x^{(t)}) \subset \mathcal{S}$. We use the Mathematical Induction to assume $\mathrm{supp}(x^{(t)}) \subset \mathcal{S}$ and consider $x^{(t+1)}$. For $i \notin \mathcal{S}$, i.e., $(x^\star)_i = 0$. If $(x^{(t+1)})_i \neq 0$ , note that $y = Ax^\star$, there is
\begin{equation}
\label{b_1}
\begin{aligned}
b^{(t)} &< |(x^{(t+1)})_i|\\
&<|{[(I-U^{(t)}A)x^{(t)} + U^{(t)}Ax^\star]}_i|\\
&=|{[(I-U^{(t)}A)(x^{(t)} - x^\star) + x^\star]}_i|\\
&\leq |{[(I-U^{(t)}A)(x^{(t)} - x^\star)]}_i| + |(x^\star)_i| \\
&= |{[(I-U^{(t)}A)(x^{(t)} - x^\star)]}_i|\\
&=|\sum_{j}(I-U^{(t)}A)_{ij}(x^{(t)} - x^\star)_j|\\
&\leq \sum_{j}|(I-U^{(t)}A)_{ij}(x^{(t)} - x^\star)_j|\\
&\leq \sum_{j} \mu(A)|(x^{(t)} - x^\star)_j|\\
&\leq \mu(A)\|x^{(t)}-x^\star\|_1.
\end{aligned}
\end{equation}
From above derivation, we can also conclude that $$|{[(I-U^{(t)}A)(x^{(t)} - x^\star)]}_i| \leq \mu(A)\|x^{(t)}-x^\star\|_1.$$ 
Since  $\mathrm{supp}(x^{(t)}) \subset \mathcal{S}$, we have $|\mathrm{supp}(x^{(t)}-x^\star)|\leq |\mathcal{S}|$, there is $$\|(I-U^{(t)}A)(x^{(t)}-x^\star)\|_1 \leq |\mathcal{S}|\mu(A)\|x^{(t)}-x^\star\|_1.$$
Since $\|U^{(t)}A(x^{(t)}-x^\star)\|_1 = \|(x^{(t)}-x^\star)-(I-U^{(t)}A)(x^{(t)}-x^\star)\|_1$, we have
\begin{equation}
\label{QAXs}
\begin{aligned}
(1-|\mathcal{S}|\mu(A)) \|x^{(t)}-x^\star\|_1 \leq &\|U^{(t)}A(x^{(t)}-x^\star)\|_1 \leq (1+|\mathcal{S}|\mu(A)) \|x^{(t)}-x^\star\|_1.
\end{aligned}
\end{equation}
As $b^{(t)}=\rho^{(t)}\|U^{(t)}(Ax^{(t)}-y)\|_1$ and $\rho^{(t)} = \frac{\mu(A)}{1-\mu(A)s} \geq \frac{\mu(A)}{1-\mu(A)|\mathcal{S}|}$, there is 
\begin{equation}
\label{rho}
b^{(t)} = \rho^{(t)}\|U^{(t)}A(x^{(t)}-x^\star)\|_1 \geq \mu(A)\|x^{(t)}-x^\star\|_1.
\end{equation}
Eq.~(\ref{b_1}) and (\ref{rho}) are conflicted, which means $(x^{(t+1)})_i = 0$ if $(x^\star)_i = 0$, i.e. $\mathrm{supp}(x^{(t+1)}) \subset \mathcal{S}$. Note that $x^{(0)} = 0 \subset \mathcal{S}$, due to the Mathematical Induction, the no "false positive" property has been proved.

Due to the inequality $x-b \leq \mathrm{sh}_{b}(x) \leq x+b$, we have $|\mathrm{sh}_{b}(x)-\zeta|\leq |x-\zeta| + |b|$, $\forall \zeta$. Thus, when we consider the absolute value of the $i$-th element of $x^{(t+1)} - x^\star$, from Eq.~(\ref{thresh_2}), we have
\begin{equation}
\label{delta1i}
\begin{aligned}
|(x^{(t+1)} - x^\star)_i| &= |(\mathrm{sh}_{b^{(t)}}((I-U^{(t)}A)x^{(t)}+U^{(t)}y) - x^\star)_i|\\
&\leq |(((I-U^{(t)}A)x^{(t)}+U^{(t)}y) - x^\star)_i| + |b^{(t)}|\\
&=|(((I-U^{(t)}A)x^{(t)}+U^{(t)}Ax^\star) - x^\star)_i| + |b^{(t)}|\\
&\leq |((I-U^{(t)}A)(x^{(t)} - x^\star))_i| + |b^{(t)}|.
\end{aligned}
\end{equation}
Since $\mathrm{supp}(x^{(t+1)}) \subset \mathcal{S}$, we have $\|x^{(t+1)}-x^\star\|_1 = \sum_{i\in\mathcal{S}}|(x^{(t+1)}-x^\star)_i|$. There we have
\begin{equation}
\label{delta2}
\begin{aligned}
\|x^{(t+1)}-x^\star\|_1 &\leq \sum_{i\in\mathcal{S}}(|((I-U^{(t)}A)(x^{(t)} - x^\star))_i| + |b^{(t)}|) \\
&= \sum_{i\in\mathcal{S}}(|\sum_{j\in\mathcal{S}\setminus\{i\}}(I-U^{(t)}A)_{ij}(x^{(t)} - x^\star)_j| + |b^{(t)}|)\\
&\leq \sum_{i\in\mathcal{S}}\sum_{j\in\mathcal{S}\setminus\{i\}}|(I-U^{(t)}A)_{ij}(x^{(t)} - x^\star)_j| + |\mathcal{S}||b^{(t)}|\\
&\leq (|\mathcal{S}|-1)\mu(A)\|x^{(t)}-x^\star\|_1 + |\mathcal{S}|\rho^{(t)}\|U^{(t)}A(x^{(t)}-x^\star)\|_1\\
&\leq (|\mathcal{S}|-1)\mu(A)\|x^{(t)}-x^\star\|_1 + \frac{\mu(A)|\mathcal{S}|}{1-\mu(A)s}\|U^{(t)}A(x^{(t)}-x^\star)\|_1.\\
&\leq (|\mathcal{S}| + |\mathcal{S}|\frac{1+\mu(A)s}{1-\mu(A)s} - 1)\mu(A)\|(x^{(t)}-x^\star)\|_1.
\end{aligned}
\end{equation}
The final step holds because $|\mathcal{S}| \leq s$ and Eq.(\ref{QAXs}).
The $l_2$ error bound of t-th output of EBT-LISTA can be calculated as
\begin{equation}
\label{t12}
\begin{aligned}
\|x^{(t)}-x^\star\|_2 &\leq \|x^{(t)}-x^\star\|_1\\
&\leq ((|\mathcal{S}| + |\mathcal{S}|\frac{1+\mu(A)s}{1-\mu(A)s} - 1)\mu(A))^t\|(x^{(0)}-x^\star)\|_1\\
&\leq q_0\exp(c_1t),
\end{aligned}
\end{equation}
where $q_0=\|x^\star\|_1$, and $c_1 = \log((|\mathcal{S}| + |\mathcal{S}|\frac{1+\mu(A)s}{1-\mu(A)s} - 1)\mu(A))$.
Compare $c_1$ with $c_0$, we have
\begin{equation}
\label{c}
\exp(c_0) - \exp(c_1) = 2\mu(A)(s - \frac{|\mathcal{S}|}{1-\mu(A)s}) >0
\end{equation}
hold when $|\mathcal{S}| < s(1-\mu(A)s)$. Under this circumstance, we have
\begin{equation}
\label{q}
q_0=\|x^\star\|_1 \leq |\mathcal{S}|B < s(1-\mu(A)s) B \leq sB.
\end{equation}
Note that $x^\star$ is sampled from $\gamma(B,s)$, Eq.~(\ref{c}) and (\ref{q}) hold with the probability with of $1-\eta$, where
\begin{equation}
\label{eta}
\begin{aligned}
\eta &= \frac{s - |\mathcal{S}|}{s} \\
&= \mu(A)s.
\end{aligned}
\end{equation}

\subsection{Proof of Lemma 2}
Remind that the update rule of LISTA with support selection is formulated as $x^{(0)}=0$, and, for $t=0,\dots,d$, 
\begin{equation}
\label{listacpss}
x^{(t+1)}=\mathrm{shp}_{(b^{(t)},p^{(t)})}((I-U^{(t)}A)x^{(t)}+U^{(t)}y),
\end{equation}

\setcounter{Lemma}{1}
\begin{Lemma}
	\label{lemma2}
	\rm{\textbf{(Convergence of LISTA with support selection)}}
	For LISTA with support selection formulated in Eq.~(\ref{listacpss}), $x^\star$ is sampled from $\gamma(B,s)$. If $s$ is small such that $\mu(A)(2s-1)<1$, $U^{(t)} \in \mathcal{W}(A)$ and $b^{(t)} = \mu(A)\sup_{x^\star}\|x^{(t)}-x^\star\|_1$,
	there actually exist two convergence phases.
	
	In the first phase, i.e., $t\leq t_0$, the $t$-th layer estimation $x^{(t)}$ satisfies $$\|x^{(t)}-x^\star\|_2\leq sB\exp{(c_2t)},$$ where $c_2 \leq \log((2s-1)\mu(A))$.
	In the second phase, i.e., $t> t_0$, the estimation $x^{(t)}$ satisfies $$\|x^{(t)}-x^\star\|_2\leq C \|x^{(t-1)}-x^\star\|_2,$$ where $C \leq s\mu(A)$. 
\end{Lemma}

\textbf{Proof.}
We want to stress that the proving process of Lemma 2 is inspired by Theorem 3 in~\cite{chen2018theoretical}.
First, we have 
\begin{equation}
\label{delta1_ss}
\begin{aligned}
(x^{(t+1)} - x^\star)_i =& \mathrm{shp}_{(b^{(t)},p^{(t)})}((I-U^{(t)}A)x^{(t)}+U^{(t)}y)_i - x^\star_i\\
=&\left\{
\begin{aligned}
    &\mathrm{sh}_{b^{(t)}}((I-U^{(t)}A)x^{(t)}+U^{(t)}y)_i - x^\star_i,\quad i\in S_{p^{(t+1)}},\\
    &((I-U^{(t)}A)x^{(t)}+U^{(t)}y)_i - x^\star_i,\quad i\notin S_{p^{(t+1)}}.\\
\end{aligned}
\right.\\
\end{aligned}
\end{equation}


Where $S_{p^{(t+1)}}$ is the set of the index of the largest $p\%$ elements (in absolute value) in vector $x^{(t+1)}$. We let $g^{(t)}_i = 0$, when $i\in S_{p^(t)}$, and $g^{(t)}_i= 1$ otherwise.
Similar to Eq.~(\ref{delta1i}), there is
\begin{equation}
\label{delta1i_ss}
\begin{aligned}
 |(x^{(t+1)} - x^\star)_i| &\leq |(I-U^{(t)}A)(x^{(t)} - x^\star)_i| + |b^{(t)}\odot g^{(t+1)}_i|.
\end{aligned}
\end{equation}
Since $b^{(t)} = \mu(A)\sup_{x^\star}\|x^{(t)}-x^\star\|_1$, same as the standard LISTA, LISTA with support selection is also "no false positive"~\cite{chen2018theoretical}, i.e., $\mathrm{supp}(x^{(t+1)}) \subset \mathcal{S}$. Therefore, we have $\|x^{(t+1)}-x^\star\|_1 = \sum_{i\in\mathcal{S}}(x^{(t+1)}-x^\star)_i$. Similar to Eq.(\ref{delta2}), we have
\begin{equation}
\label{delta2_ss}
\begin{aligned}
\|x^{(t+1)}-x^\star\|_1 \leq& \sum_{i\in\mathcal{S}}(|(I-U^{(t)}A)(x^{(t)} - x^\star)_i| + |b^{(t)}\odot g^{(t+1)}_i|) \\
=& \sum_{i\in\mathcal{S}}(|\sum_{j\in\mathcal{S}\setminus\{i\}}(I-U^{(t)}A)_{ij}(x^{(t)} - x^\star)_j|+ |b^{(t)}\odot g^{(t+1)}_i|)\\
\leq& \sum_{i\in\mathcal{S}}\sum_{j\in\mathcal{S}\setminus\{i\}}|(I-U^{(t)}A)_{ij}(x^{(t)} - x^\star)_j| + \sum_{i\in\mathcal{S}}|b^{(t)}\odot g^{(t+1)}_i|\\
\leq& (|\mathcal{S}|-1)\mu(A)\|x^{(t)}-x^\star\|_1 + \sum_{i\in\mathcal{S}}|b^{(t)}\odot g^{(t+1)}_i|.\\
\end{aligned}
\end{equation}
We let $S_{t+1}$ denote the number of non-zero entries in $x^{(t+1)}$. Also, $P_{t+1}$ denotes the number of the largest $p^{(t+1)}\%$ elements (in absolute value) in $x^{(t+1)}$. Therefore the number of zero entries in $g(x^{(t+1)})$ is $\min(S_{t+1},P_{t+1})$. Then Eq.(\ref{delta2_ss}) can be calculated as
\begin{equation}
\label{delta2_ss2}
\begin{aligned}
\|x^{(t+1)}-x^\star\|_1 \leq& (|\mathcal{S}|-1)\mu(A)\|x^{(t)}-x^\star\|_1 + \sum_{i\in\mathcal{S}}|b^{(t)}\odot g(x^{(t+1)}_i)|\\
\leq& (|\mathcal{S}|-1)\mu(A)\|x^{(t)}-x^\star\|_1 + (|\mathcal{S}|-\min(S_{t+1},P_{t+1}))|b^{(t)}|\\
=&(|\mathcal{S}|-1)\mu(A)\|x^{(t)}-x^\star\|_1 + (|\mathcal{S}|-\min(S_{t+1},P_{t+1}))\mu(A)\sup_{x^\star}\|x^{(t)}-x^\star\|_1.\\
\end{aligned}
\end{equation}
We now take the supremum of Eq.(\ref{delta2_ss2}), there is
\begin{equation}
\label{delta_sup_ss}
\begin{aligned}
\sup_{x^\star}\|x^{(t+1)}-x^\star\|_1 \leq(2s-1-\min(S_{t+1},P_{t+1}))\mu(A)\sup_{x^\star}\|x^{(t)}-x^\star\|_1.
\end{aligned}
\end{equation}
Note that $\|x^\star\|_1 \leq sB$. Assume $k = \arg\min_{t}(S_t,P_t)$, the $l_2$ upper bound of t-th output can be calculated as
\begin{equation}
\label{t11_ss}
\begin{aligned}
\|x^{(t)}-x^\star\|_2&\leq \|x^{(t)}-x^\star\|_1 \leq \sup_{x^\star}\|x^{(t)}-x^\star\|_1\\
&\leq \left(\prod_{i=1}^{t}(2s-1-\min(S_{i},P_{i}))\mu(A)\right)\sup_{x^\star}\|x^{(0)}-x^\star\|_1\\
&\leq ((2s-1-\min(S_{k},P_{k}))\mu(A))^{{t}}sB\\
&\leq sB\exp(c_2t),
\end{aligned}
\end{equation}
where $c_2 = \log((2s-1-\min(S_{k},P_{k}))\mu(A))$. Apparently, we have $c_2 \leq c_0 = \log((2s-1)\mu(A))$.

From Eq.(\ref{t11_ss}), we have $\|x^{(t)}-x^\star\|_1 \leq sB\exp(c_2t)$, which means $l_1$ error bound can approaches to 0. Thus, there exists a $t^*$, when $t>t^*$, $\|x^{(t)}-x^\star\|_1 \leq min_{i\in \mathcal{S}}(x^\star)_i$. Note that $|x^{(t)}_i - (x^\star)_i| \leq \|x^{(t)}-x^\star\|_1$. If $i \in \mathcal{S}$, i.e., $(x^\star)_i \neq 0$, there exists $x^{(t)}_i \neq 0$, which means $\mathcal{S} \subset \mathrm{supp}(x^{(t)})$. Recall the "no false positive" property, i.e., $\mathrm{supp}(x^{(t)}) \subset \mathcal{S}$, we can conclude that $\mathrm{supp}(x^{(t)}) = \mathcal{S}$. Since $P_{t}$ increases layerwise and $P_{max}$ is set as the upper bound of $|\mathcal{S}|$, there exists $t'$ statisfies $P_{t'} > s$, we let $t_0=\max(t^*,t')$, if $t > t_0$, there exists $P_{t} \geq |\mathcal{S}|$ and $\mathrm{supp}(x^{(t)}) = \mathcal{S}$. Under this circumstance, if $i \in \mathcal{S}$, we have $x^{(t)}_i \neq 0$ and $i \in S_{p_{t}}$, which means every element in $\mathcal{S}$ will be selected as support. There we have
\begin{equation}
\label{xt+1_s2}
\begin{aligned}
x^{(t+1)}_i - (x^\star)_i=& \mathrm{shp}_{(b^{(t)},p^{(t)})}(((I-U^{(t)}A)x^{(t)}+U^{(t)}Ax^\star)_i) - (x^\star)_i\\
=& ((I-U^{(t)}A)x^{(t)}+U^{(t)}Ax^\star)_i - (x^\star)_i\\
=& ((I-U^{(t)}A)(x^{(t)}-x^\star))_i.
\end{aligned}
\end{equation}
We let $x_{\mathcal{S}} \in \mathbb{R}^{|\mathcal{S}|}$ denote the vector that keeps the elements with indices of $x$ in $\mathcal{S}$ and removes the others. Similarly, we let $M(\mathcal{S},\mathcal{S}) \in \mathbb{R}^{|\mathcal{S}|\times |\mathcal{S}|}$ denote the submatrix of matrix $M$ which keeps the row and column if the index belongs to $ \mathcal{S}$. Then, we have
\begin{equation}
\label{delta1_e}
\begin{aligned}
\|x^{(t+1)}-x^\star\|_2 &= \|(x^{(t+1)}-x^\star)_{\mathcal{S}}\|_2\\
&= \|((I-U^{(t)}A)(x^{(t)}-x^\star))_{\mathcal{S}}\|_2\\
&= \|(I-U^{(t)}A)(\mathcal{S},\mathcal{S})(x^{(t)}-x^\star)_{\mathcal{S}}\|_2\\
&\leq \|(I-U^{(t)}A)(\mathcal{S},\mathcal{S})\|_2 \|(x^{(t)}-x^\star)_{\mathcal{S}}\|_2\\
&=  C\|(x^{(t)}-x^\star)\|_2,
\end{aligned}
\end{equation}
where $C = \|(I-U^{(t)}A)(\mathcal{S},\mathcal{S})\|_2$. Further we have $C \leq \|(I-U^{(t)}A)(\mathcal{S},\mathcal{S})\|_F \leq \sqrt{|\mathcal{S}|^2\mu(A)^2} \leq s\mu(A)$. 
\subsection{Proof of Theorem 2}
Remind that our EBT-LISTA with support selection can be formulated as $x^{(0)}=0$ and and for $t=0,\dots,d$,
\begin{equation}
\label{eq2}
\begin{aligned}
x^{(t+1)}&=\mathrm{shp}_{(b^{(t)},p^{(t)})}((I-U^{(t)}A)x^{(t)}+U^{(t)}y),\\
b^{(t)}&=\rho^{(t)}\|U^{(t)}(Ax^{(t)}-y)\|_\phi.
\end{aligned}
\end{equation}

\begin{Theorem}
	\label{thm2}
	\rm{\textbf{(Convergence of EBT-LISTA with support selection)}}
	For EBT-LISTA with support selection and $\phi=1$, $x^\star$ is sampled from $\gamma(B,s)$. If $s$ is small such that $\mu(A)(2s-1) < 1$, $U^{(t)} \in \mathcal{W}(A)$, and $\rho^{(t)} = \frac{\mu(A)}{1-\mu(A)s}$,
	there exist two convergence phases.
	
	In the first phase, i.e., $t\leq t_1$, the $t$-th layer estimation $x^{(t)}$ satisfies $$\|x^{(t)}-x^\star\|_2\leq q_1\exp{(c_3t)},$$ where $c_3<c_2$, $q_1 < sB$ and $t_1<t_0$ hold with a probability of $1 - \mu(A)s$.
	In the second phase, i.e., $t> t_1$, the estimation $x^{(t)}$ satisfies $$\|x^{(t)}-x^\star\|_2\leq C \|x^{(t-1)}-x^\star\|_2,$$ where $C \leq s\mu(A)$. 
\end{Theorem}

\textbf{Proof.} From Eq.~(\ref{eq2}), similar to Eq.~(\ref{delta1_ss}) and (\ref{delta1i_ss}), we have
\begin{equation}
\label{delta2i_ss}
\begin{aligned}
|(x^{(t+1)} - x^\star)_i| \leq |(I-U^{(t)}A)(x^{(t)} - x^\star)_i| + |b^{(t)}\odot g^{(t+1)}_i|,
\end{aligned}
\end{equation}
where $b^{(t)} = \rho^{(t)}\|U^{(t)}(Ax^{(t)}-y)\|_1 = \frac{\mu(A)}{1-\mu(A)s}\|U^{(t)}A(x^{(t)}-x^\star)\|_1$. Same as the origin EBT-LISTA, EBT-LISTA with support selection is also "no false positive" and Eq.~(\ref{QAXs}) hold either. Therefore $\|U^{(t)}A(x^{(t)}-x^\star)\|_1 \leq (1+|\mathcal{S}|\mu(A)) \|x^{(t)}-x^\star\|_1 \leq (1+s\mu(A)) \|x^{(t)}-x^\star\|_1$. Similar to Eq.~(\ref{delta2_ss}) and (\ref{delta2_ss2}), there is
\begin{equation}
\label{delta3_ss}
\begin{aligned}
\|x^{(t+1)}-x^\star\|_1 =& \sum_{i\in\mathcal{S}}|(x^{(t+1)} - x^\star)_i|\\
\leq& \sum_{i\in\mathcal{S}}(|(I-U^{(t)}A)(x^{(t)} - x^\star)_i| + |b^{(t)}\odot g(x^{(t+1)}_i)|) \\
=& \sum_{i\in\mathcal{S}}(|\sum_{j\in\mathcal{S}\setminus\{i\}}(I-U^{(t)}A)_{ij}(x^{(t)} - x^\star)_j| +|b^{(t)}\odot g^{(t+1)}_i\\
\leq& \sum_{i\in\mathcal{S}}\sum_{j\in\mathcal{S}\setminus\{i\}}|(I-U^{(t)}A)_{ij}(x^{(t)} - x^\star)_j| + \sum_{i\in\mathcal{S}}|b^{(t)}\odot g^{(t+1)}_i|\\
\leq& (|\mathcal{S}|-1)\mu(A)\|x^{(t)}-x^\star\|_1 + (|\mathcal{S}|-\min(S_{t+1},P_{t+1}))|b^{(t)}|\\
\leq& (|\mathcal{S}|-1)\mu(A)\|x^{(t)}-x^\star\|_1 + (|\mathcal{S}|-\min(S_{t+1},P_{t+1}))\mu(A)\frac{1+\mu(A)s}{1-\mu(A)s}\|(x^{(t)}-x^\star)\|_1\\
\leq& (\frac{2}{1-\mu(A)s}|\mathcal{S}|-\frac{1+\mu(A)s}{1-\mu(A)s}\min(S_{t+1},P_{t+1})-1)\mu(A)\|(x^{(t)}-x^\star)\|_1.
\end{aligned}
\end{equation}
Similar to Eq.~(\ref{t11_ss}), the $l_2$ error bound can be calculated as
\begin{equation}
\label{t22}
\begin{aligned}
\|x^{(t)}-x^\star\|_2\leq& \|x^{(t)}-x^\star\|_1\\
\leq& \prod_{i=1}^{t}\left[(\frac{2}{1-\mu(A)s}|\mathcal{S}|-\frac{1+\mu(A)s}{1-\mu(A)s}\min(S_{i},P_{i})-1)\mu(A)\right]\|(x^{(0)}-x^\star)\|_1\\
\leq&\left[(\frac{2}{1-\mu(A)s}|\mathcal{S}|-\frac{1+\mu(A)s}{1-\mu(A)s}\min(S_{k},P_{k})-1)\mu(A)\right]^t\|x^\star\|_1\\
\leq& q_1\exp(c_3t),
\end{aligned}
\end{equation}
where we have $q_1 = \|x^\star\|_1$, and $c_3 = \log((\frac{2}{1-\mu(A)s}|\mathcal{S}|-\frac{1+\mu(A)s}{1-\mu(A)s}\min(S_{k},P_{k})-1)\mu(A))$, and .

Compare $c_3$ with $c_2$, we have
\begin{equation}
\label{c2}
\begin{aligned}
\exp(c_2) - \exp(c_3)&= 2\mu(A)(s - \frac{|\mathcal{S}|}{1-\mu(A)s} + \frac{2\mu(A)s}{1-\mu(A)s}\min(S_{k},P_{k})) \\
&\geq 2\mu(A)(s - \frac{|\mathcal{S}|}{1-\mu(A)s})>0\\
\end{aligned}
\end{equation}
hold when $|\mathcal{S}| < s(1-\mu(A)s)$. Under this circumstance, we have
\begin{equation}
\label{q2}
q_1=\|x^\star\|_1 \leq |\mathcal{S}B < s(1-\mu(A)s) B \leq sB.
\end{equation}
Note that $x^\star$ is sampled from $\gamma(B,s)$, Eq.~(\ref{c2}) and (\ref{q2}) hold with the probability with of $1-\eta$, where
\begin{equation}
\label{eta2}
\begin{aligned}
\eta &= \frac{s - |\mathcal{S}|}{s} \\
&= \mu(A)s.
\end{aligned}
\end{equation}
Similar to LISTA with support selection, there exists a $t^{**}$, when $t>t^{**}$, $\|x^{(t)}-x^\star\|_1 \leq min_{i\in \mathcal{S}}(x^\star)_i$. Therefore, $\mathrm{supp}(x^{(t)}) = \mathcal{S}$. Recall that $c_3 < c_2$ holds with the probability of $1-\eta$, we have $t^{**} < t^{*}$ with the probability of $1-\eta$. When we use the same settings for $p^{(t)}$ as LISTA-SS, then we have same $t'$ satisfying $P_{t'} > s$. Let $t_1=\max(t^{**},t')$, we have $t_1 \leq t_0$ with the probability of $1-\eta$. When $t > t_1$, we have $x^{(t)}_i \neq 0$ and $i \in S_{p_{t}}$.  
Same as Eq.~(\ref{xt+1_s2}) and (\ref{delta1_e}), there is
\begin{equation}
\label{delta2_e}
\|x^{(t+1)}-x^\star\|_2 \leq C\|(x^{(t)}-x^\star)\|_2,
\end{equation}
where $C = \|(I-U^{(t)}A)(\mathcal{S},\mathcal{S})\|_2 \leq \|(I-U^{(t)}A)(\mathcal{S},\mathcal{S})\|_F \leq \sqrt{|\mathcal{S}|^2\mu(A)^2} \leq s\mu(A)$.

\bibliographystyle{IEEEbib}
\bibliography{refs}